%% file: 01_main.tex
\documentclass[10pt,twocolumn,letterpaper]{article}

\usepackage[pagenumbers]{cvpr} 

\input{preambles/packages}
\input{preambles/macro}

\definecolor{cvprblue}{rgb}{0.21,0.49,0.74}
\usepackage[pagebackref,breaklinks,colorlinks,allcolors=cvprblue]{hyperref}

\def\paperID{40169} 
\def\confName{CVPR}
\def\confYear{2026}

\title{Improving Text-to-Image Generation with Intrinsic Self-Confidence Rewards}

\author{Seungwook Kim$^{1,3}$ \hspace{2.0cm} Minsu Cho$^{1,2}$ \vspace{1.0mm} \\
$^1$POSTECH \hspace{1.5cm} $^2$RLWRLD \hspace{1.5cm} $^3$GenGenAI \\
\small
\href{https://wookiekim.github.io/SOLACE/}{\url{https://wookiekim.github.io/SOLACE/}}
}
\vspace{-2.0mm}

\begin{document}


\twocolumn[{
\renewcommand\twocolumn[1][]{#1}%
\maketitle
\vspace{-8.0mm}
\includegraphics[width=1.0\linewidth]{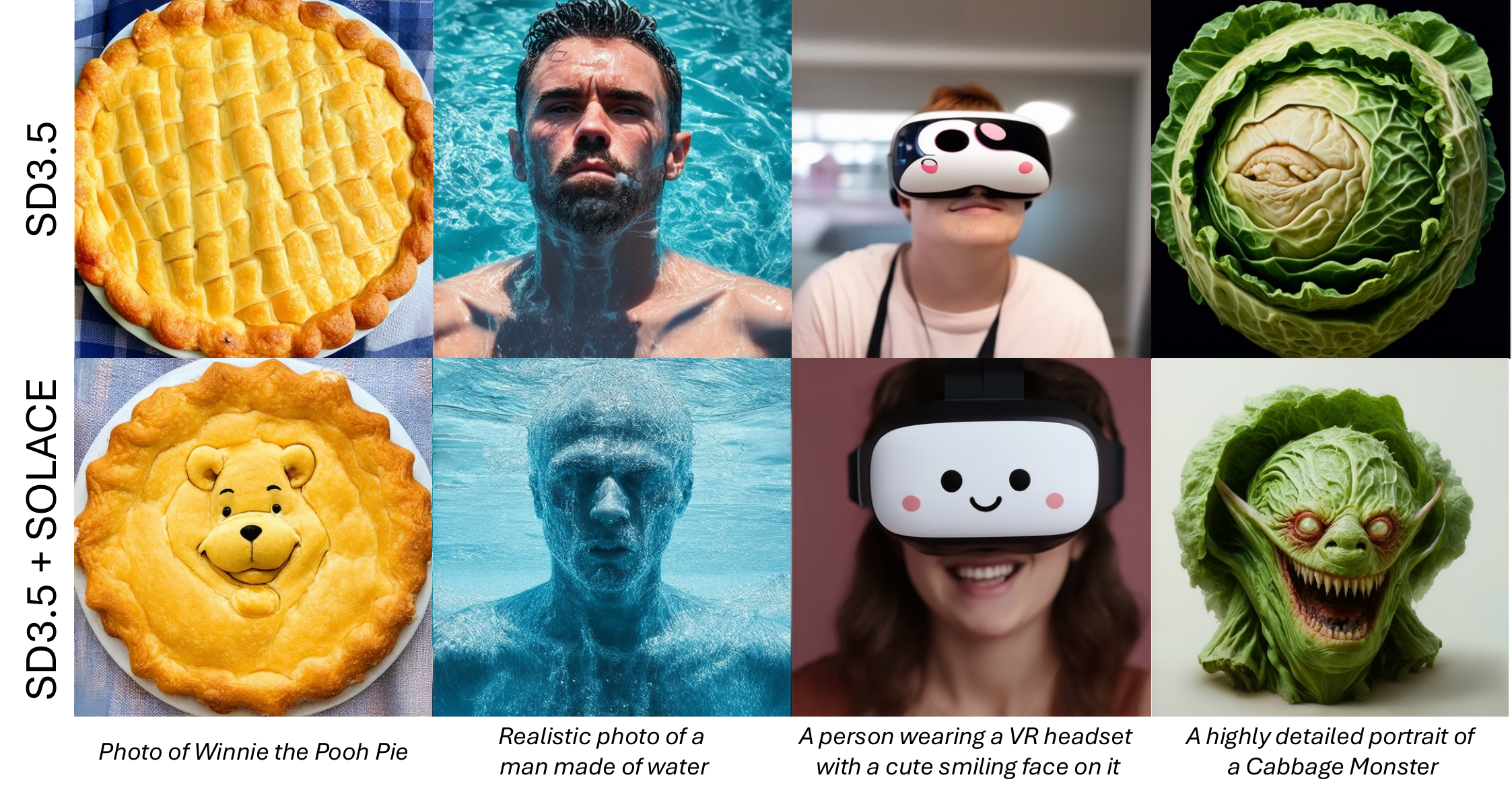}
\vspace{-8.0mm}
\captionof{figure}{\textbf{Qualitative examples of SOLACE on Pick-a-Pic dataset~\cite{kirstain2023pick}.}
Best viewed on electronics.
} 
\vspace{+2.0mm}
\label{fig:teaser}
}]

\input{sections/0_abstract}

\input{sections/1_introduction}

\input{sections/2_relatedwork}

\input{sections/3_preliminary}

\input{sections/4_method}

\input{sections/5_experiments}

\input{sections/6_conclusion}

\clearpage
\clearpage

\clearpage
\noindent
\small
\textbf{Acknowledgement.} 
This work was supported by the IITP grants (RS-2022-II220290: Visual Intelligence for Space-Time Understanding and Generation based on Multi-layered Visual Common Sense (40\%), RS-2022-II220113: Developing a Sustainable Collaborative Multi-modal Lifelong Learning Framework (50\%), RS-2019-II191906: AI Graduate School Program at POSTECH (5\%), RS-2025-02653113: High-Performance Research AI Computing Infrastructure Support at the 2 PFLOPS Scale (5\%)) funded by the Korea government (MSIT).
This work was also supported by the Scaleup TIPS grant (RS-2023-00321784: Development of Novel Generative AI Technology to Generate Domain-Specific Synthetic Data).

{
    \small
    \bibliographystyle{ieeenat_fullname}
    \bibliography{main}
}

\input{sections/X_appendix}

\end{document}

%% file: preambles/packages.tex
\usepackage{graphicx}
\usepackage{amsmath}
\usepackage{amssymb}
\usepackage{booktabs}
\usepackage{dsfont}
\usepackage{multirow}
\usepackage{graphics}
\usepackage{wrapfig}
\usepackage{bbm}
\usepackage{xspace}
\usepackage{xcolor}
\usepackage{tikz}
\usepackage{pifont}
\usepackage{multirow}
\usepackage[table]{xcolor}
\usepackage{subcaption} 
\usepackage{caption} 
\usepackage{tabularx}

%% file: preambles/macro.tex
\usepackage{xr}
\makeatletter
\newcommand*{\addFileDependency}[1]{
  \typeout{(#1)}
  \@addtofilelist{#1}
  \IfFileExists{#1}{}{\typeout{No file #1.}}
}
\makeatother


%% file: sections/0_abstract.tex
\begin{abstract}

Text-to-image generation powers content creation across design, media, and data augmentation.
Post-training of text-to-image generative models is a promising path to improve human preference alignment, factuality, and aesthetics.
We introduce SOLACE (\textbf{S}elf-\textbf{O}riginating \textbf{LA}tent \textbf{C}onfidence \textbf{E}stimation), a post-training framework that replaces external reward supervision with an internal \emph{self-confidence} signal: we re-noise the model's own outputs and measure how accurately it recovers the injected noise, treating low reconstruction error as high self-confidence.
SOLACE converts this intrinsic signal into scalar rewards for reinforcement learning, requiring no external reward models, annotators, or preference data.
By reinforcing high-confidence generations, SOLACE delivers consistent gains in compositional generation, text rendering, and text-image alignment.
Integrating SOLACE with external rewards yields complementary improvements while alleviating reward hacking.

\end{abstract}

%% file: sections/1_introduction.tex
\section{Introduction}
\label{sec:intro}


Text-to-image (T2I) generation has advanced rapidly with the rise of diffusion and flow-based models, delivering high-fidelity, diverse images from natural language prompts~\cite{rombach2022high, saharia2022photorealistic, podell2023sdxl, peebles2023scalable, chen2023pixart, chen2024pixart, esser2024scaling,gong2025seedream}.
These models now support a broad range of applications: controllable image editing and inpainting~\cite{brooks2023instructpix2pix, boesel2024improving, shi2024seededit,  batifol2025flux, sheynin2024emu, xiao2025omnigen, zhang2025context}; serving as powerful priors or pre-trained components for text-to-video diffusion models~\cite{zhou2022magicvideo, wang2023modelscope, guo2023animatediff, hacohen2024ltx, kong2024hunyuanvideo, wan2025wan, kim2025freeaction}; data creation and augmentation pipelines for downstream perception tasks~\cite{shin2023fill, yang2023ai, wang2024domain}; and text-to-3D (and 4D) reconstruction via score distillation sampling~\cite{poole2022dreamfusion, shi2023mvdream, wang2023prolificdreamer, singer2023text, bahmani20244d, kim2024correspondentdream, kim2024multiimagedream, kim2025rapidmv}.
Recent studies show that \emph{post-training text-to-image generative models} via reinforcement learning can yield dramatic improvements in visual appeal and aesthetic quality~\cite{wallace2024diffusion, black2023training, liu2025flowgrpo}, typically by optimizing external rewards derived from human preference models~\cite{kirstain2023pick, wu2023human, xu2023imagereward} or task-specific validators~\cite{ghosh2023geneval, cui2025paddleocr}.



However, defining a scalable and reliable reward for “good” images remains challenging~\cite{kirstain2023pick,wu2023human,xu2023imagereward,lee2023holistic,sun2025t2v}.
There are numerous, weakly-aligned criteria a good image has to satisfy, \textit{e.g.,} compositionality, text rendering, aesthetics, and text–image alignment, whose relative importance shifts across domains and prompts~\cite{lee2023holistic}.
In practice, external-reward post-training is also vulnerable to over-optimization: optimizing a narrow critic can induce reward hacking and regressions on non-target capabilities, degrading coverage or faithfulness even as the targeted score rises~\cite{wallace2024diffusion,black2023training,liu2025flowgrpo}.
Human-preference based reward models~\cite{kirstain2023pick,xu2023imagereward,wang2025unified} are popular for their efficacy, but require large-scale annotation for training.
Operationally, external rewards require running additional evaluators (preference/OCR/safety models) alongside the generator during training, increasing pipeline complexity.


Despite extensive progress in extrinsically supervised post-training, intrinsic signals remain under-explored for text-to-image generation.
In this work, we ask: \textbf{\textit{can internal feedback from the text-to-image generator itself provide meaningful signals for post-training?}}
To this end, we introduce Self-Originating LAtent Confidence Estimation (SOLACE), a post-training framework that uses the model’s own \emph{self-confidence} as a reward.
Inspired by Score Distillation Sampling~\cite{poole2022dreamfusion, shi2023mvdream, wang2023prolificdreamer}, which uses a pretrained text-to-image generator as a critic for text-to-3D or -4D generation, we propose to let a text-to-image generator \textit{critique its own generation}.
Concretely, given a sampled latent $z_0$, we re-noise it to selected timesteps $t\in\mathcal{T}$ using the forward noising schedule, and measure how well the model recovers the injected noise. Low reconstruction error indicates high self-confidence.
Our hypothesis is that large-scale pretraining endows diffusion models with strong priors over real images and text-image correspondence, so self-confidence should correlate with text alignment and realism.

Empirically, SOLACE yields consistent gains in compositional generation~\cite{ghosh2023geneval}, text rendering~\cite{cui2025paddleocr}, and text-image alignment~\cite{radford2021learningclip}, while modestly improving human-preference scores~\cite{kirstain2023pick,wu2023human,xu2023imagereward,wang2025unified}, all without external rewards.
Qualitative comparisons and a user study corroborate these trends, indicating that intrinsic self-confidence aligns with key aspects of image generation quality.
Moreover, applying SOLACE on top of an \emph{extrinsically post-trained} model (\ie one already fine-tuned with external rewards) yields further improvements in compositionality, text rendering, and alignment, with only slight drops on the targeted external metric. This shows that intrinsic and extrinsic rewards are complementary, and that SOLACE alleviates the reward hacking commonly observed in external-reward post-training.


The key contributions of our work are as follows:

\begin{itemize}
\item We present \textbf{SOLACE} (\textbf{S}elf-\textbf{O}riginating \textbf{LA}tent \textbf{C}onfidence \textbf{E}stimation), a post-training framework using \emph{self-confidence} as reward.

\item We define self-confidence as the model’s ability to recover noise injected into its own outputs: we re-noise the generated latent, measure reconstruction error, and convert it into a scalar reward for GRPO post-training.

\item Across standard benchmarks and a comprehensive user study, SOLACE yields consistent gains in compositionality, text rendering, and text–image alignment, while modestly improving human-preference metrics.

\item SOLACE complements \emph{external}-reward pipelines: applying SOLACE on top of externally post-trained models improves non-target capabilities (compositionality, text rendering, alignment) with only mild trade-offs on the targeted external metric, mitigating reward hacking.

\end{itemize}

%% file: sections/2_relatedwork.tex
\section{Related Work}
\label{sec:related_work}


\smallbreak
\noindent
\textbf{Text-to-image generative models.}
Text-to-image generation is a rapidly advancing field, which was initially dominated by diffusion models~\cite{bao2023all, chen2023pixart, chen2024pixart, peebles2023scalable, podell2023sdxl, rombach2022high, saharia2022photorealistic}.
Recent work increasingly adopts flow matching~\cite{esser2024scaling, batifol2025flux, shin2025deeply} and sequence models~\cite{yu2022scaling, chang2023muse, luo2024open, sun2024autoregressive} for improved efficiency and generation quality.
Advances span architectures~\cite{peebles2023scalable, esser2024scaling, batifol2025flux}, image recaptioning~\cite{betker2023improving, chen2023pixart, chen2024pixart}, and tokenization~\cite{sun2024autoregressive, yu2024image, kim2025democratizing}.
In this work, we focus on reinforcement-learning based post-training to improve text-to-image models, using the self-confidence of the generative model as the \textit{intrinsic} reward.


\smallbreak
\noindent
\textbf{Text-to-image model alignment via post-training.}
Post-training is emerging as an effective paradigm to align existing text-to-image models toward desired objectives, \textit{e.g.,} human preference. 
This can take the form of direct fine-tuning given differentiable rewards~\cite{prabhudesai2023aligning, clark2023directly, xu2023imagereward, prabhudesai2024video} or Reward Weighted Regression (RWR)~\cite{peng2019advantage, fan2025online, lee2023aligning, dong2023raft}.
Some schemes build on reinforcement learning to leverage PPO~\cite{schulman2017proximal}-style policy gradients~\cite{black2023training, fan2023reinforcement, miao2024training, gupta2025simple, zhao2025score}, or perform Direct Preference Optimization (DPO) or its variants~\cite{rafailov2023direct, wallace2024diffusion, yuan2024self, liu2025improving, yang2024using, zhang2024onlinevpo, furuta2024improving, liang2025aesthetic, liu2025videodpo}.
More recently, Flow-GRPO~\cite{liu2025flowgrpo} introduces GRPO~\cite{shao2024deepseekmathgrpo} for flow matching models, by converting the ODE of flow matching sampling to SDEs to inject stochasticity.
However, external rewards increase training costs (an additional model must run alongside the generator) and raise the risk of reward hacking~\cite{rafailov2023direct, wallace2024diffusion, liu2025videodpo,liu2025improving}.
In this work, we define self-confidence as the model's ability to recover noise injected into its own outputs, and use this \textit{intrinsic} signal for post-training, improving compositional generation, text rendering, and text-image alignment without reward hacking.


\smallbreak
\noindent
\textbf{Intrinsic signals for post-training.}
Intrinsic signals for post-training have recently gained traction in language modeling as scalable alternatives to human-labeled preference data, leveraging self-derived feedback such as confidence/uncertainty estimates, self-evaluation, and self-consistency to guide reinforcement learning or preference optimization without annotators \cite{zelikman2022star, chen2024self, yuan2024selfllm, poesia2024learning, cheng2024self, zhao2025intuitor, xu2025genius, zuo2025ttrl, zhao2025absolute}.
Recently, Intuitor~\cite{zhao2025intuitor} showed that using self-certainty as a confidence-based intrinsic reward enables single-agent reinforcement learning across diverse tasks without relying on explicit feedback, gold labels, or environment-based validation.
Bringing the same principle to text-to-image generation is non-trivial: generation proceeds along continuous denoising trajectories and likelihoods are implicit, unlike token-level discrete objectives in LLMs. 
In this work, we define self-confidence of flow-matching models as their ability to recover noise injected into their own outputs, inspired by score-distillation sampling~\cite{poole2022dreamfusion,shi2023mvdream}.
This enables dense, on-policy feedback without labeled data or reward models. 
Empirically, we show that this self-confidence signal aligns with compositionality, text rendering, and text-image alignment.

%% file: sections/3_preliminary.tex
\section{Preliminary: GRPO for Flow Matching}
\label{sec:prelim}

\subsection{Flow Matching and Rectified Flow}

Flow matching bypasses score learning in conventional diffusion models~\cite{ho2020denoising, song2020score} by directly regressing the target velocity of a transport ODE along a user-chosen path between data and a reference distribution~\cite{lipman2022flow,liu2022flow}. 
Recent state-of-the-art generative models~\cite{esser2024scaling, batifol2025flux, wan2025wan, kong2024hunyuanvideo} adopt the Rectified Flow (RF) framework. Specifically, let $x_0\!\sim p_{\text{data}}$ and $x_1\!\sim p_1$ (e.g., $\mathcal N(0,I)$); RF chooses the straight-line path 
\begin{equation}
x_t = (1-t)\,x_0 + t\,x_1,
\end{equation}
for which the target velocity is constant in $t$:
\begin{equation}
v^\star = \partial_t x_t = x_1 - x_0 .
\end{equation}
Training reduces to direct regression of this constant velocity at random $(x_t,t)$ pairs:
\begin{equation}
\mathcal{L}(\theta)
= \mathbb{E}_{x_0\sim p_{\text{data}},\,x_1\sim p_1,\,t\sim\mathcal U[0,1]}
\Big\| v^\star - v_\theta(x_t, t)  \Big\|_2^2 .
\end{equation}
After training, sampling solves the deterministic ODE
\begin{equation}
\label{eq:rf-ode}
\frac{\mathrm{d}x_t}{\mathrm{d}t} = v_\theta(x_t,t), \qquad t:1\!\to\!0,
\end{equation}
starting from $x_1\!\sim p_1$ and transporting to $x_0$.

\input{assets/figures/main_overview}


\subsection{GRPO for Flow Matching}
\label{subsec:grpo-flow}
For a policy $\pi_\theta$, we consider a policy-gradient objective that maximizes expected cumulative reward while regularizing updates toward a reference policy $\pi_{\mathrm{ref}}$ via a KL penalty:
\begin{equation}
\label{eq:grpo-generic}
\begin{aligned}
&\max_{\theta} \mathbb{E}_{(s_0,a_0,\ldots,s_T,a_T)\sim\pi_\theta}
\\
&\Big[\sum_{t=0}^{T} R(s_t,a_t) - \beta \sum_{t=0}^{T} D_{\mathrm{KL}}\!\left(\pi_\theta(\cdot\mid s_t)\,\|\,\pi_{\mathrm{ref}}(\cdot\mid s_t)\right)
\Big],
\end{aligned}
\end{equation}
where $R(s_t,a_t)$ is the per-step reward.
Group Relative Policy Optimization (GRPO)~\cite{shao2024deepseekmathgrpo} proposes to use a group relative formulation to estimate the advantage for each sample to optimize~\cref{eq:grpo-generic}.

Flow-GRPO~\cite{liu2025flowgrpo} integrates GRPO into flow matching models for online RL post-training.
The iterative denoising process in flow matching can be formulated as a Markov Decision Process~\cite{black2023training}:
given a text prompt $c$, the flow model $p_\theta$ samples a group of $G$ images $\{x^i_0\}_{i=1}^G$ and the corresponding sampling trajectories $\{(x_T^i, x_{T-1}^i, \cdots, x_0^i)\}_{i=1}^G$. 
The advantage of the $i$-th image is calculated by normalizing the group-level rewards:
\begin{equation}
\label{eq:flow_grpo_adv}
\hat{A}^i_t = \frac{R(x_0^i,c) - \textrm{mean}(\{R(x_0^i,c)\}_{i=1}^G)}{\textrm{std}(\{R(x_0^i,c)\}_{i=1}^G)}
\end{equation}
Finally, GRPO optimizes the policy model by maximizing $\mathcal{J}_\textrm{Flow-GRPO} = \mathbb{E}_{c\sim C, \{x^i\}_{i=1}^G\sim \pi_{\theta_\textrm{old}}(\cdot | c)}f(r,\hat{A},\theta,\epsilon,\beta)$, where

\begin{equation}
\label{eq:grpo-f-short}
\begin{aligned}
f(r,\widehat{A},\theta,\epsilon,\beta)
&= \underset{i,t}{\mathrm{mean}}
\Big[
\min\!\big(r_t^{\,i},\,\mathrm{clip}_\epsilon(r_t^{\,i})\big)\,\widehat{A}_t^{\,i}
\Big]
\;-\;\beta\,\overline{D}_{\mathrm{KL}}, \\
\overline{D}_{\mathrm{KL}}
&= \underset{t}{\mathrm{mean}}\;
D_{\mathrm{KL}}\!\left(\pi_\theta(\cdot\mid s_t)\,\|\,\pi_{\mathrm{ref}}(\cdot\mid s_t)\right),
\\
\mathrm{clip}_\epsilon(r)\;&\triangleq\;\mathrm{clip}(r,\,1-\epsilon,\,1+\epsilon).
\end{aligned}
\end{equation}
and $r^i_t(\theta) = \frac{p_\theta(x^i_{t-1} | x^i_T,c)}{p_{\theta_\textrm{old}}(x^i_{t-1} | x^i_T,c)}$.
Flow-GRPO then converts the deterministic ODE of~\cref{eq:rf-ode} into an equivalent SDE that matches the original model's marginal probability function at all timesteps, in order to meet the GRPO policy update requirements, \textit{e.g.}, stochasticity is necessary for exploration in RL post-training.
We adopt Flow-GRPO to post-train flow-matching text-to-image models.


%% file: assets/figures/main_overview.tex
\begin{figure*}[t]
\begin{center}
\includegraphics[width=1.0\linewidth]{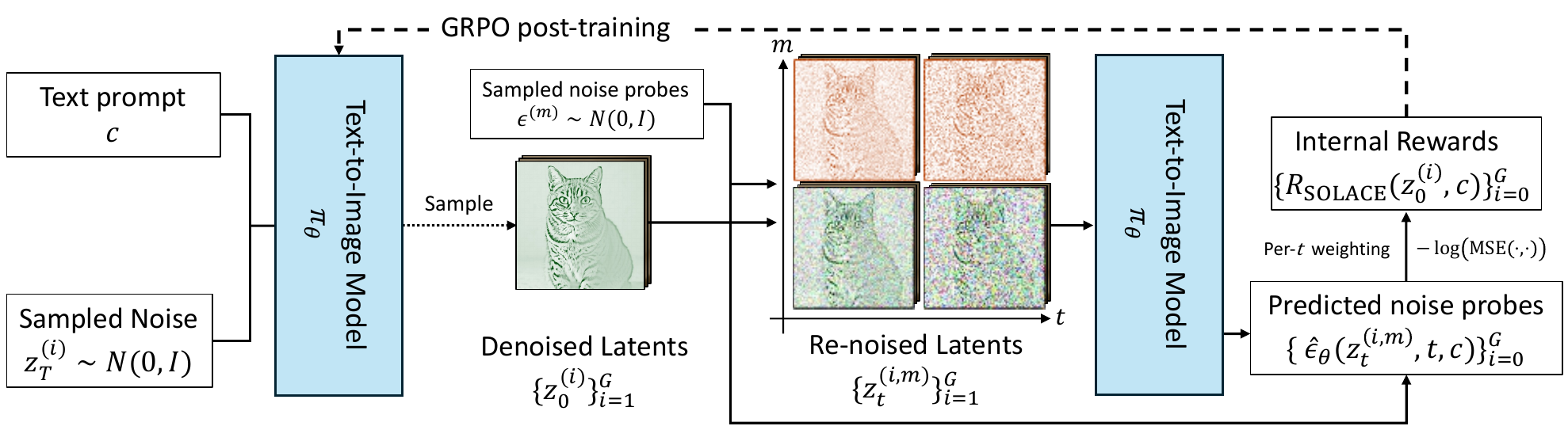}
\end{center}
\vspace{-5.0mm}
\caption{\textbf{Overview of SOLACE.} 
Given a text prompt $c$, we generate $G$ different latents. Without decoding, we re-noise the latents using $K$ noise probes across $t\in\mathcal{T}\subset[0,1]$.
For each generated latent $z_0^{(i)}$, we formulate the text-to-image generative model's self-confidence of the generated latent as the ability to denoise the re-noised latent. 
We leverage this self-confidence as an internal reward scalar value, which we use to post-train the text-to-image generative model using GRPO~\cite{shao2024deepseekmathgrpo, liu2025flowgrpo}. We omit the KL term in this figure for better readability.
}
\vspace{-5.0mm}
\label{fig:main_overview}
\end{figure*}

%% file: sections/4_method.tex
\section{Method: SOLACE}
\label{sec:method}


\smallbreak
\noindent
\textbf{Overview.}
We present \textbf{SOLACE} (Self-Originating LAtent Confidence Estimation), a post-training method for text-to-image generators that requires no external reward models.
SOLACE uses the model’s own \emph{self-confidence} as an intrinsic reward: after generating an output, we re-noise it at selected timesteps and measure how accurately the model recovers the injected noise.
Aggregating these per-timestep recovery errors yields a single on-policy scalar reward for reinforcement learning.
In the following, we detail the computation of the self-confidence reward (\cref{subsec:method_sdsr}) and the stabilization techniques for SOLACE training (\cref{subsec:method_technical_considerations}).
An overview of SOLACE is shown in \cref{fig:main_overview}.

\subsection{Intrinsic Self-Confidence Reward}
\label{subsec:method_sdsr}

\smallbreak
\noindent
\textbf{Sampling a group of images for GRPO.} 
Given a text prompt $c$, we sample $G$ independent reverse trajectories in the latent space $\mathcal Z$ under the flow policy $\pi_\theta$:
\begin{equation}
\label{eq:rollout-latent}
z_T^{(i)} \sim \mathcal{N}(0, I),
z_{t-1}^{(i)} \sim \pi_\theta\!\left(\cdot \mid z_t^{(i)},\, c\right),i=1,\ldots,G.
\end{equation}
This produces terminal latents $\{z_0^{(i)}\}_{i=1}^G$ and trajectories $\{(z_T^{(i)},z_{T-1}^{(i)},\ldots,z_0^{(i)})\}_{i=1}^G$.
Using multiple independent draws yields the group required for group-relative advantage normalization in GRPO.
While we can sample $G$ different images from the same initial noise $z_T$ due to the added stochasticity from~\cite{liu2025flowgrpo}, we sample different initial noise to improve exploration during GRPO training.

\smallbreak
\noindent
\textbf{Sampling noise probes for re-noising.}
We draw a shared set of $K$ noise probes in latent space:
\begin{equation}
\label{eq:probe-sampling-shared}
\epsilon^{(m)} \sim \mathcal{N}(0, I), \qquad m=1,\ldots,K,
\end{equation}
so that candidate $i$ and candidate $j$ are perturbed by the \emph{same} probes $\{\epsilon^{(m)}\}_{m=1}^K$.
For rectified flow, we re-noise a terminal latent $z_0^{(i)}$ via the linear forward kernel
\begin{equation}
\label{eq:rf-forward}
z_t^{(i,m)} \;=\; (1-t)\, z_0^{(i)} \;+\; t\, \epsilon^{(m)}, 
\qquad t \in \mathcal{T}\subset[0,1],
\end{equation}
where $\mathcal{T}$ is the set of re-noising levels used for evaluation. 
We take $K$ even ($K\!\ge\!2$) and use antithetic pairing to enforce exact mean zero within the probe set, i.e., $\epsilon^{(m+K/2)}=-\,\epsilon^{(m)}$ for $m=1,\ldots,K/2$.

\smallbreak
\noindent
\textbf{Calculating self-confidence.}
For each noised latent $z_t^{(i,m)}$ (Eq.~\eqref{eq:rf-forward}), we query the flow-matching model’s velocity field $v_\theta(z_t^{(i,m)},t,c)$.
Under the rectified-flow parameterization, the velocity predicts a linear transform of the injected noise; specifically, we recover a noise estimate via
\begin{equation}
\label{eq:eps-hat}
\widehat{\epsilon}_\theta\!\left(z_t^{(i,m)},t,c\right)\;=\;v_\theta\!\left(z_t^{(i,m)},t,c\right)\;+\;z_0^{(i)}.
\end{equation}
We then measure the reconstruction error against $\epsilon^{(m)}$:
\begin{equation}
\label{eq:mse-probe}
\mathrm{MSE}_{i,t}
\;=\;
\frac{1}{K}\sum_{m=1}^{K}
\left\|
\widehat{\epsilon}_\theta\!\left(z_t^{(i,m)},t,c\right)-\epsilon^{(m)}
\right\|_2^2.
\end{equation}
To turn small errors into large rewards while stabilizing dynamic range, we use the negative log transform,
\begin{equation}
\label{eq:sdsr-step}
S_{i,t}
\;=\;
-\,\log\!\big(\mathrm{MSE}_{i,t}+\delta\big),
\end{equation}
where $\delta>0$ avoids $\log 0$. This choice (i) approximates a Gaussian log-likelihood score under an i.i.d.\ noise model, (ii) compresses outliers, and (iii) yields additive contributions across timesteps.
Aggregating over a set of re-noising levels $\mathcal{T}\subset[0,1]$ gives the scalar intrinsic reward
\begin{equation}
\label{eq:sdsr-scalar}
R_{\mathrm{SOLACE}} \big(z_0^{(i)},c\big)
\;=\;
\frac{1}{\sum_{t\in\mathcal{T}} w(t)}
\sum_{t\in\mathcal{T}} w(t)\, S_{i,t}.
\end{equation}
We use $w(t)=1$ in practice for simplicity.
Note that external rewards typically operate in pixel space, $R_{\mathrm{ext}}(x^{(i)},c)$, where $x^{(i)}=\mathrm{Dec}(z_0^{(i)})$ for a fixed decoder $\mathrm{Dec}\!:\mathcal Z\!\to\!\mathcal X$. 
In contrast, $R_{\mathrm{SOLACE}}$ is computed \emph{directly in latent space}, avoiding decoding and keeping the signal model-native.

\begin{figure*}[htbp]
\begin{center}
\includegraphics[width=1.0\linewidth]{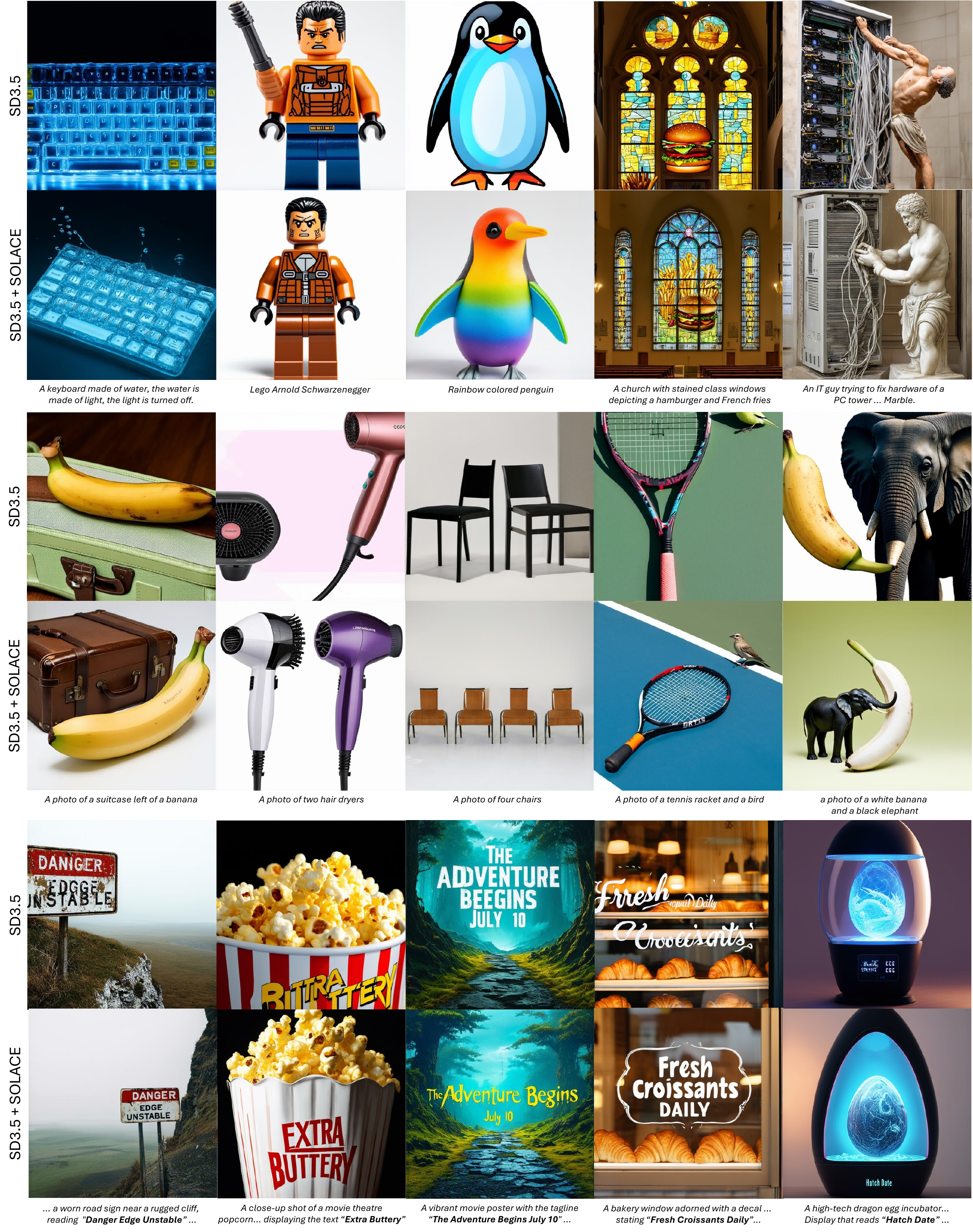}
\end{center}
\vspace{-5mm}
\caption{\textbf{Qualitative results of SOLACE on SD3.5~\cite{esser2024scaling} across DrawBench~\cite{saharia2022photorealistic}, GenEval~\cite{ghosh2023geneval} and OCR~\cite{cui2025paddleocr}.} SOLACE shows consistent improvements over the baseline SD3.5.}
\label{fig:main_qualitative}
\end{figure*}

\subsection{Stabilization and Efficiency Techniques}
\label{subsec:method_technical_considerations}


\smallbreak
\noindent
\textbf{Denoising reduction for efficient training.}
Following Flow-GRPO~\cite{liu2025flowgrpo}, we shorten the reverse-time horizon by subsampling the denoising steps.
This reduces compute without degrading gains: e.g., while SD3.5 uses $40$ steps at inference, we use 10 during training.
We find that this does not sacrifice image quality at test time, while enabling faster training.

\smallbreak
\noindent
\textbf{Timestep selection for self-confidence probing.}
We probe self-confidence at the \emph{exact scheduler timesteps} used by the SD3.5 sampler (same discretization and indices), ensuring alignment with the generation trajectory.
This avoids mismatch between sampling and probing, yielding more reliable credit assignment.

\smallbreak
\noindent
\textbf{Training on selective timesteps.}
We observe that training on all denoising timesteps easily leads to collapse (e.g., blank or textureless images), a form of reward hacking in which the model steers latents toward regimes where injected noise becomes trivially easy to predict.
We mitigate this by training on only a suffix of the schedule, \ie a fixed percentage of the later reverse steps, where the denoising task remains informative but is harder to exploit.  
Let $\mathcal{T}_{\mathrm{train}}\!\subset\!\mathcal{T}$ denote this suffix window ($\lvert\mathcal{T}_{\mathrm{train}}\rvert=\lceil \rho\,\lvert\mathcal{T}\rvert\rceil$); we apply GRPO losses only on $t\!\in\!\mathcal{T}_{\mathrm{train}}$, which stabilizes learning without collapse.

\smallbreak
\noindent
\textbf{CFG-free self-confidence computation.}
Although $G$ images are sampled with CFG for GRPO training, SOLACE self-confidence is computed \emph{without} CFG.
CFG forms a mixture field $v_{\mathrm{cfg}} = v_{\mathrm{uncond}} + s\,(v_{\mathrm{cond}}-v_{\mathrm{uncond}})$; computing self-confidence on this mixture would measure confidence of the guided proxy rather than the base conditional model.
Empirically, omitting CFG during self-confidence computation yields stronger and more stable improvements.

\smallbreak
\noindent
\textbf{Online calculation of self-confidence.}
We can compute self-confidence either (1) online, using the model being trained ($\pi_\theta$), or (2) offline, using a fixed base model ($\pi_\textrm{ref}$).
While offline computation does not cause severe over-optimization~\cite{gao2023scaling}, online computation yields better performance.
We conjecture that as the model improves through SOLACE post-training, its self-confidence estimates become more reliable, reinforcing further gains.

\input{assets/tables/main_quantitative}


%% file: assets/tables/main_quantitative.tex
\begin{figure*}[ht]

\centering
\small
\begin{tabular}{l c  c c c c c c c}
\toprule
 &  \multicolumn{2}{c}{Task-specific} & \multicolumn{2}{c}{Image Quality} & \multicolumn{4}{c}{Human Preference} \\
\cmidrule(lr){2-3}\cmidrule(lr){4-5} \cmidrule(lr){6-9}
Model & GenEval & OCR  & ClipScore & Aesthetic & PickScore & HPSv2.1 & ImageReward & UnifiedReward\\
\midrule
{\color{gray}SDXL}   & {\color{gray}0.55} & {\color{gray}0.14} & {\color{gray}0.287} & {\color{gray}5.60} & {\color{gray}22.42} & {\color{gray}0.280} & {\color{gray}0.76} & {\color{gray}2.93} \\

{\color{gray}SD3.5-L } & {\color{gray}0.71} & {\color{gray}0.68} &   {\color{gray}0.289} & {\color{gray}5.50} & {\color{gray}22.91} & {\color{gray}0.288} & {\color{gray}0.96} & {\color{gray}3.25} \\

\midrule



SD3.5-M                & 0.65 & 0.61 & 0.282 & 5.36 & 22.34 & 0.279  & 0.84  & 3.08 \\

\textbf{+ SOLACE} (Ours)              & \textbf{0.71} & \textbf{0.67} & \textbf{0.288} & \textbf{5.39} & \textbf{22.41} & 0.278 & \textbf{0.87} & \textbf{3.11} \\

\midrule
\midrule

\multirow{3}{*}{\shortstack{SD3.5-M\\+ FlowGRPO}}
& \cellcolor{blue!18}0.95 &  0.65  & 0.293  & 5.32 & 22.51& 0.272 & 1.06 & 3.18 \\

    & 0.67 & \cellcolor{blue!18}0.92  & 0.290 & 5.32 & 22.41 & 0.280 & 0.95 & 3.14 \\
    
    & 0.54 & 0.68 & 0.278 & 5.90 & \cellcolor{blue!18}23.50 & 0.314 & 1.26 & 3.37 \\
\midrule

\multirow{3}{*}{\shortstack{SD3.5-M\\+ FlowGRPO \\ +\textbf{SOLACE} (Ours)}}

    & \cellcolor{blue!18}0.92 & \textbf{0.71} & \textbf{0.294} & \textbf{5.35} & 22.50 & \textbf{0.277} & 1.06 & \textbf{3.26} \\

    & \textbf{0.72} & \cellcolor{blue!18}0.89  & \textbf{0.291} & \textbf{5.39} & \textbf{22.45} & \textbf{0.284} & \textbf{0.97} & \textbf{3.19} \\
    
    & \textbf{0.77} & \textbf{0.70} &  \textbf{0.287} & 5.63 & \cellcolor{blue!18}22.73 & 0.286 & 1.07 & 3.26 \\
\bottomrule

\bottomrule
\end{tabular}
\captionof{table}{
\textbf{Quantitative results of SOLACE.}
We evaluate SOLACE on SD3.5~\cite{esser2024scaling} across GenEval~\cite{ghosh2023geneval}, Text Rendering, human preference models~\cite{kirstain2023pick,wu2023human,xu2023imagereward,wang2025unified}, and image quality metrics.
SOLACE yields consistent gains across all quantitative metrics.
In the bottom section, each row of SD3.5-M + FlowGRPO corresponds to a different external reward used for FlowGRPO training; the \colorbox{blue!18}{blue cell} indicates which metric was used as the external reward.
}
\label{tab:main_quantitative}

\includegraphics[width=1.0\linewidth]{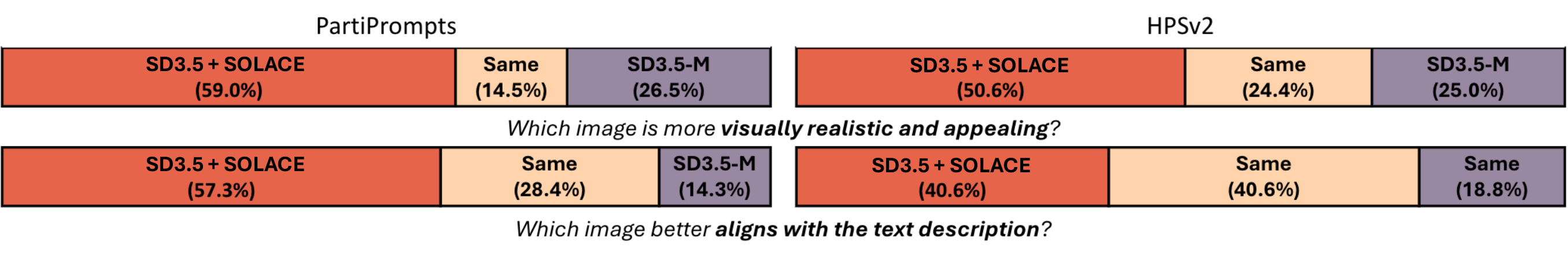}
\vspace{-9.0mm}
\captionof{figure}{\textbf{User study against baseline SD3.5-M~\cite{esser2024scaling} on PartiPrompts~\cite{saharia2022photorealistic} and HPSv2~\cite{wu2023human}.}
The user study shows that SOLACE post-training yields favorable visual realism/appeal, and text-image alignment.}
\label{fig:user_study}
\vspace{-2.0mm}
\end{figure*}

%% file: sections/5_experiments.tex
\section{Experiments}
\label{sec:experiments}

\subsection{Implementation details}
\smallbreak
\noindent
We use a group size $G=16$ and number of noise probes $K=8$ with antithetic pairing in our experiments.
While SOLACE requires no external reward models, annotators, or preference data for training, it does require a \emph{prompt corpus} to generate the terminal latents for training; we use the train set of the visual text rendering task~\cite{cui2025paddleocr} from Flow-GRPO~\cite{liu2025flowgrpo}, which holds longer and more informative prompts compared to Pick-a-Pic~\cite{kirstain2023pick} or GenEval~\cite{ghosh2023geneval}. We note that SOLACE improves across different prompt sources (see supplementary \cref{appendix:ablation_study}).
We use LoRA~\cite{hu2022lora, peft} with rank $r=32$ and scaling factor $\alpha=64$ for parameter-efficient post-training.
We use the AdamW~\cite{loshchilov2017decoupled} optimizer with constant learning rate of 3e$^{-4}$, and the KL regularizer weight $\beta = 0.04$.
In $\lvert\mathcal{T}_{\mathrm{train}}\rvert=\lceil \rho\,\lvert\mathcal{T}\rvert\rceil$, we set $\rho=0.6$, which yields improvements without reward hacking or training collapse.
An image resolution of 512$\times$512 is used for both training and testing.
We use a CFG guidance scale of 7.0 at inference.
All experiments are carried out on $8\times$NVIDIA RTX PRO 6000 Blackwell GPUs.
We include more training details in the supplementary materials.


\subsection{Evaluation setting}

\smallbreak
\noindent
\textbf{(1) Compositional image generation.} We evaluate on GenEval~\cite{ghosh2023geneval}, consisting of complex compositional prompts including object counting, attribute binding, and spatial relations.
Evaluation is performed across six tasks: position, counting, attribute binding, colors, two objects, and single object. 
We follow the official evaluation pipeline, which detects object bounding boxes and colors, then infers spatial relations from the generated image.
The scores are calculated in a rule-based manner \eg for object counting, $r = 1 - \frac{| N_\textrm{gen} - N_\textrm{ref}|}{N_\textrm{ref}}$, where $N_\textrm{gen}$ is the number of generated objects, while $N_\textrm{ref}$ is the specified number of objects in the prompt.

\smallbreak
\noindent
\textbf{(2) Visual text rendering.} We use the 1,000 GPT4o~\cite{gpt4o}-generated test prompts from~\cite{liu2025flowgrpo}.
In each prompt, the exact string that should appear in the image (\ie target text) is specified by \texttt{"\{text\}"}.
We adhere to~\cite{gong2025seedream} to report $r = \textrm{max}(0, 1-\frac{N_e}{N_\textrm{ref}})$, where $N_e$ is the minimum edit distance between the rendered text and the target text, and $N_\textrm{ref}$ is the non-whitespace length of the target text.

\smallbreak
\noindent
\textbf{(3) Human preference alignment.} We report the model-based reward outputs from Pickscore~\cite{kirstain2023pick}, HPSv2~\cite{wu2023human}, ImageReward~\cite{xu2023imagereward} and UnifiedReward~\cite{wang2025unified}, trained on large-scale human preference data. 
We use the test prompts from DrawBench~\cite{saharia2022photorealistic} to generate the images for evaluation.

\smallbreak
\noindent
\textbf{(4) Image quality evaluation.} We additionally report the CLIP-Score~\cite{radford2021learningclip} and Aesthetic Score~\cite{schuhmann2022laion} on DrawBench~\cite{saharia2022photorealistic}, to evaluate the overall quality of generated images independent of the above task-specific criteria. 



\subsection{Results}

\smallbreak
\noindent
\textbf{Quantitative results.}
Results are shown in \cref{tab:main_quantitative}. Applying SOLACE on SD3.5-M yields consistent gains across task-specific, image quality, and human preference metrics.
While improvements in human preference are modest, we observe substantial gains in compositional generation (GenEval~\cite{ghosh2023geneval}), text rendering (OCR~\cite{cui2025paddleocr}), and CLIPScore~\cite{radford2021learningclip}, nearly matching the performance of SD3.5-L in these metrics despite having less than $\frac{1}{3}$ of the parameters (2.5B vs.\ 8.1B).
This shows that the model's intrinsic self-confidence is strongly correlated with compositionality, text rendering, and text-image alignment.

\input{assets/figures/flowgrpo_arc}

\input{assets/tables/main_ablation}
We also analyze the effect of applying SOLACE \textit{after} post-training SD3.5-M with external rewards via Flow-GRPO~\cite{liu2025flowgrpo}.
The results show that while performance on the targeted external reward is mildly compromised, we consistently gain improvements across GenEval, OCR, and CLIPScore. This strengthens our hypothesis that intrinsic self-confidence is strongly correlated with compositionality, text rendering, and text-image alignment, and that SOLACE alleviates the reward hacking typically seen in external-reward post-training.
In~\cref{fig:flowgrpo_solace}, we show visual examples of SD3.5-M post-trained with FlowGRPO (PickScore), then further post-trained with SOLACE, showing that the two rewards are complementary.

\smallbreak
\noindent
\textbf{User study.}
In~\cref{fig:user_study}, we provide the results of a user study on prompts from PartiPrompt~\cite{yu2022scaling} and HPSv2~\cite{wu2023human}, asking users to assess the generated images based on visual appeal/realism and text alignment.
We summarize $\sim$3,600 responses from 40 participants.
The results show that SD3.5-M post-trained with SOLACE consistently outperforms the baseline in both visual realism/appeal and text alignment.

\smallbreak
\noindent
\textbf{Qualitative comparison.}
We provide additional qualitative comparisons in~\cref{fig:teaser} and~\cref{fig:main_qualitative}, showing that SOLACE yields visually appealing results with improved compositionality and text rendering, even without any external reward.
We note that SOLACE learns to generate images more tailored to the given prompt; when prompted with detailed descriptions, SOLACE produces realistic outputs.





\subsection{Ablation study and analyses}
\label{sec:ablation}

In~\cref{tab:main_ablation}, we provide ablation study results to validate the design and hyperparameter choices of SOLACE.
\smallbreak
\noindent
\textbf{Analyses on number of noise probes $K$.}
We vary $K$ across $4, 8, 16$. The results show that $K=8$ yields slightly better results overall.
While $K=16$ slightly outperforms $K=8$ in aesthetic score, the improvement is negligible relative to the additional compute cost.



\smallbreak
\noindent
\textbf{CFG for self-confidence.}
Using CFG during self-confidence computation results in a slight performance drop.
We conjecture this is because CFG is an inference-time technique, and using it inside the reward would optimize the \emph{guided proxy} rather than the base conditional policy $\pi_\theta(\cdot\mid z_t,c)$.
This may incentivize reward hacking via guidance strength rather than learning a better $\pi_\theta$.

\smallbreak
\noindent
\textbf{Online self-confidence vs Offline self-confidence.}
We compare post-training performance when self-confidence is computed online (\ie using the model being trained, $\pi_\theta$) versus offline (\ie using the fixed base model, $\pi_\textrm{ref}$).
Using offline self-confidence as a static reward results in lower performance across metrics, suggesting that online computation, which improves alongside the model, provides a stronger training signal.



\smallbreak
\noindent
\textbf{Observed causes of training collapse.}
Training collapses when (1) we train on too many timesteps, \ie $\rho > 0.6$ in $\lvert\mathcal{T}_{\mathrm{train}}\rvert=\lceil \rho\,\lvert\mathcal{T}\rvert\rceil$, or (2) we do not use CFG for sampling the $G$ candidates.
In both cases, over-optimization against the self-confidence reward occurs, producing textureless images due to reward hacking. See the supplementary for detailed analysis.

\begin{figure}[h]
\begin{center}
\includegraphics[width=1.0\linewidth]{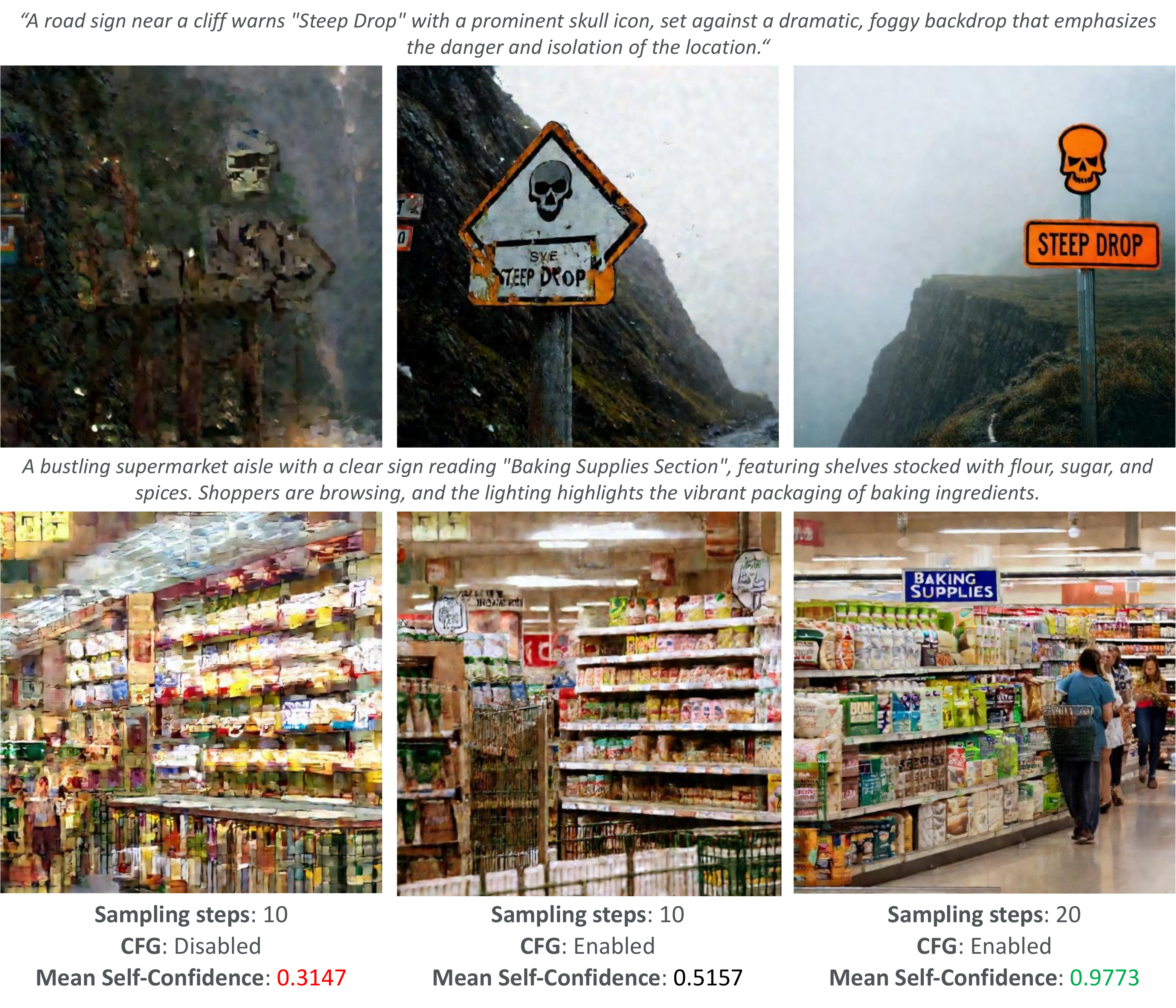}
\end{center}
\vspace{-6mm}
\caption{\textbf{Rationale of SOLACE.} Distributions of self-confidence under three inference settings. The distribution shifts rightward (higher self-confidence) as visual quality improves, showing that noise recovery accuracy is predictive of sample quality.}
\vspace{-6mm}
\label{fig:solace_rationale}
\end{figure}

\smallbreak
\noindent
\textbf{Rationale of self-confidence as reward.}
We test whether self-confidence correlates with image quality by comparing three inference regimes: (i) $10$ steps without CFG, (ii) $10$ steps with CFG, and (iii) $20$ steps with CFG.
As shown in~\cref{fig:solace_rationale}, the self-confidence distribution shifts rightward from (i) to (iii), matching the rise in visual quality. Since the \emph{same} model computes the signal regardless of guidance or step count, better samples are easier to self-denoise, motivating self-confidence as a reward.

\subsection{Limitations of SOLACE}
One limitation is that intrinsic self-confidence does not align strongly with human preference; observed gains on preference metrics are modest.
Also, while SOLACE improves compositional generation, text rendering, and text faithfulness, it cannot target a specific alignment objective on its own.
However, we showed that SOLACE can be integrated with external rewards to target specific alignments while alleviating reward hacking and improving compositionality or text rendering capabilities (\cref{tab:main_quantitative}).
We note that SOLACE's self-confidence is computed \emph{under the same text conditioning} $c$ in $r(x,c)$, which reduces (but does not eliminate) the risk of reinforcing prompt-agnostic high-density modes; we provide empirical analysis on rare compositions and diversity preservation in the supplementary (\cref{appendix:diversity_semantic}).


%% file: assets/figures/flowgrpo_arc.tex
\begin{figure}[htbp]
\begin{center}
\includegraphics[width=1.00\linewidth]{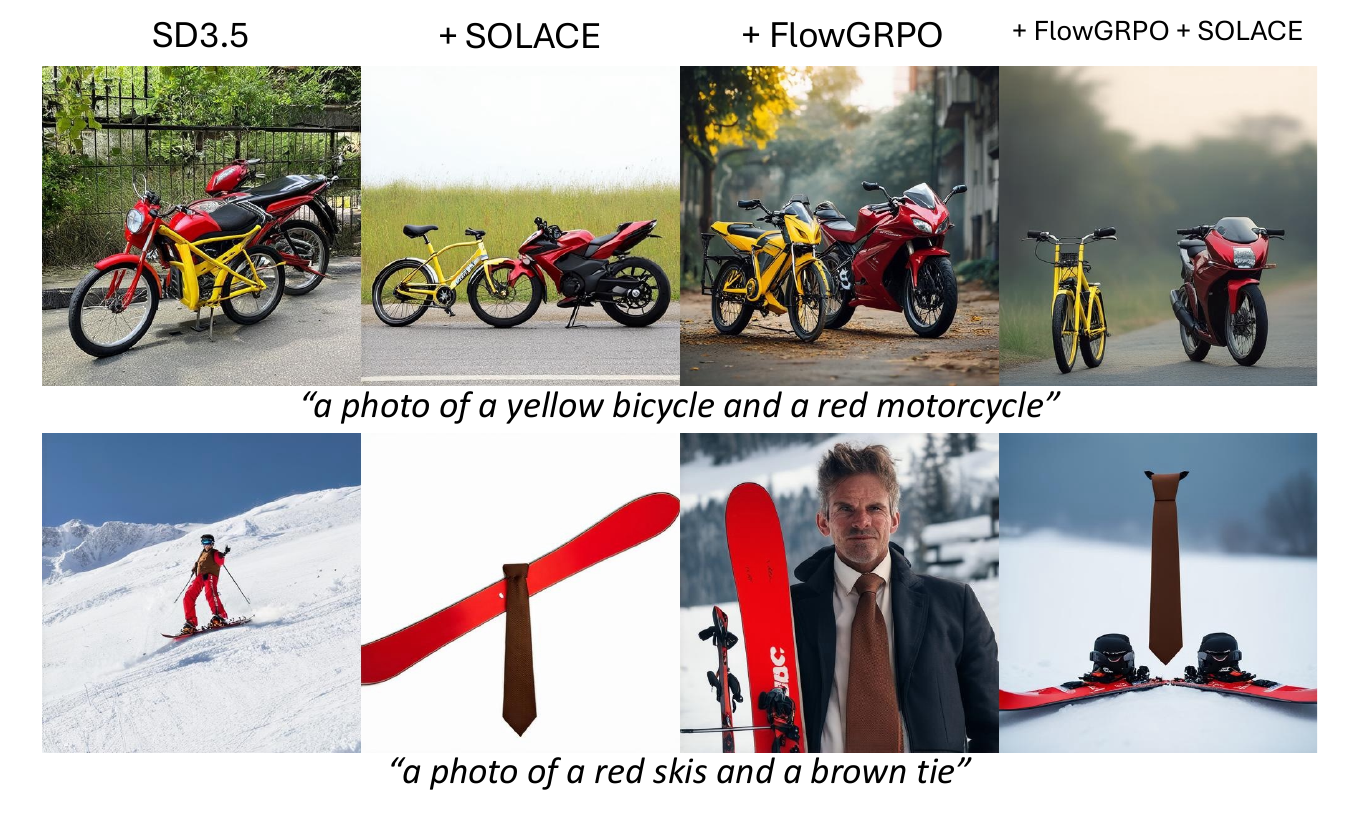}
\end{center}
\vspace{-5.0mm}
\caption{\textbf{Effect of SOLACE post-training SD3.5-M after post-training on PickScore~\cite{kirstain2023pick} using FlowGRPO~\cite{liu2025flowgrpo}.} 
SOLACE complements external rewards, showing the best compositional generation and visual appeal on GenEval~\cite{ghosh2023geneval}.
Post-training on external rewards yields high visual appeal, but sacrifices compositionality as shown above (Column 3: Generates yellow motorcycle instead / generates unwanted human).}
\label{fig:flowgrpo_solace}
\vspace{-6.0mm}
\end{figure}

%% file: assets/tables/main_ablation.tex
\begin{table*}[!ht]

\centering
\small
\begin{tabular}{l c c c c c c c c}
\toprule
 & \multicolumn{2}{c}{Task-specific} & \multicolumn{2}{c}{Image Quality} & \multicolumn{4}{c}{Human Preference} \\
\cmidrule(lr){2-3}\cmidrule(lr){4-5} \cmidrule(lr){6-9}

  & GenEval & OCR  & ClipScore & Aesthetic & PickScore & HPSv2.1 & ImageReward & UnifiedReward\\
\midrule
\multicolumn{9}{l}{\textit{Number of noise probes $K$}} \\
\midrule
$K=4$ & 0.71 & 0.66 &  0.287 & 5.37 & 22.34 & 0.273 & 0.81 & 3.08 \\
$K=8$ (Ours) & 0.71 & 0.67 & 0.288 & 5.39 & 22.41 & 0.278 & 0.87 & 3.11 \\
$K=16$ & 0.70 & 0.67 & 0.288 & 5.42 & 22.34 & 0.278 & 0.86 & 3.09 \\ 





\midrule
\multicolumn{9}{l}{\textit{Classifier-Free Guidance for self-confidence calculation}} \\
\midrule
O & 0.68 & 0.59 & 0.287 & 5.38 & 22.39 & 0.278 & 0.85 & 3.10 \\
X (Ours) & 0.71 & 0.67 & 0.288 & 5.39 & 22.41 & 0.278 & 0.87 & 3.11 \\

\midrule
\multicolumn{9}{l}{\textit{Offline vs Online Self-Confidence}} \\
\midrule
Offline & 0.69 & 0.61 & 0.285 & 5.36 & 22.36 & 0.274 & 0.82 & 3.07 \\
Online (Ours) & 0.71 & 0.67 & 0.288 & 5.39 & 22.41 & 0.278 & 0.87 & 3.11 \\

\bottomrule
\end{tabular}
\vspace{-1.0mm}
\caption{\textbf{Ablation study results of SOLACE.} We validate the design choices of SOLACE over number of noise probes $K$, the usage of CFG for self-confidence calculation, and online/offline self-confidence calculation.
Our current configurations yield superior results.}
\label{tab:main_ablation}
\vspace{-6.0mm}
\end{table*}

%% file: sections/6_conclusion.tex
\section{Conclusion}
\label{sec:conclusion}

We introduced SOLACE, a post-training framework that replaces external rewards with intrinsic self-confidence, defined as the model's ability to recover noise injected into its own outputs.
Across benchmarks and a user study, reinforcing higher self-confidence yields consistent improvements in compositionality, text rendering, and text-image alignment.
SOLACE also complements external rewards: applying it on externally post-trained models improves non-target capabilities while alleviating reward hacking.
We demonstrate SOLACE's generality across architectures, model scales, resolutions, and modalities in the supplementary.
Future directions include (i) multi-view extensions to carry SOLACE to 3D and 4D generation, and (ii) calibrating intrinsic signals for task-targeted reward shaping.

%% file: sections/X_appendix.tex
\clearpage
\setcounter{page}{1}
\maketitlesupplementary


\section*{Supplementary Contents}
\pdfbookmark[1]{Supplementary Contents}{supp-contents}

\begin{itemize}
  \item \hyperref[appendix:solace_sdlarge]{SOLACE Post-Training on SD3.5-L}
   \item \hyperref[appendix:flux_solace]{Applying SOLACE on FLUX.1-Dev}
   \item \hyperref[appendix:solace_sdxl]{Applying SOLACE on SDXL}
   \item \hyperref[appendix:t2v]{SOLACE for Text-to-Video Generation}
   \item \hyperref[appendix:resolution]{Resolution Analysis}
  \item \hyperref[appendix:closed_source]{Comparison with Closed-Source Models}
   \item \hyperref[appendix:training_collapse_analysis]{Training Collapse Analysis}
  \item \hyperref[appendix:diversity_semantic]{Diversity and Semantic Correctness Analysis}
  \item \hyperref[appendix:positive_only]{Effect of Negative Advantages}
  \item \hyperref[appendix:ablation_study]{Additional Ablation Studies}
  \item \hyperref[appendix:implementation_details]{Additional Implementation Details}
  \item \hyperref[appendix:user_study_interface]{User Study Instructions and Interface}
  \item \hyperref[appendix:additional_qual]{Additional Qualitative Results}
\end{itemize}


\section{SOLACE Post-Training on SD3.5-L}
\label{appendix:solace_sdlarge}

To assess scalability, we apply SOLACE to SD3.5-L~\cite{esser2024scaling}, a larger base model than the SD3.5-M used in the main experiments.
Unless otherwise noted, we reuse the same training recipe (shortened denoising horizon, suffix-only updates, shared probes, CFG-free scoring).
As reported in \cref{tab:supp_other_models}, SOLACE yields \emph{consistent gains} in compositional generation, text rendering, and text–image alignment, while remaining competitive on human-preference metrics (e.g., HPSv2, PickScore).
These results suggest that SOLACE scales to higher-capacity text-to-image models without inducing reward hacking and remains effective beyond the SD3.5-M setting.






\section{Applying SOLACE on FLUX.1-Dev}
\label{appendix:flux_solace}

To test architectural generality, we apply SOLACE to \textbf{FLUX.1\,-\,Dev}~\cite{batifol2025flux}, a flow-matching text-to-image generator with a design distinct from SD3.5.
We keep the core SOLACE recipe unchanged (shortened denoising horizon, suffix-only updates, shared probes, CFG-free scoring), adapting only to the model's native scheduler and inference step count.
A small deviation is the suffix window: we set $\rho=0.5$, i.e., train on the latter half of the scheduler steps, which increased training stability in this setting.
As reported in \cref{tab:supp_other_models}, SOLACE delivers \emph{consistent gains} in compositional generation, text rendering, and text–image alignment, while remaining competitive on  human-preference metrics (e.g., HPSv2, PickScore).
The results indicate that SOLACE transfers effectively across architectures and remains robust on another representative flow-matching T2I model.

\begin{table*}[!ht]

\centering
\small
\begin{tabular}{l c c c c c c c c}
\toprule
 & \multicolumn{2}{c}{Task-specific} & \multicolumn{2}{c}{Image Quality} & \multicolumn{4}{c}{Human Preference} \\
\cmidrule(lr){2-3}\cmidrule(lr){4-5} \cmidrule(lr){6-9}

  & GenEval & OCR  & ClipScore & Aesthetic & PickScore & HPSv2.1 & ImageReward & UnifiedReward\\
\midrule
SD3.5-M (2.5B)               & 0.65 & 0.61 & 0.282 & 5.36 & 22.34 & 0.279  & 0.84  & 3.08 \\

\textbf{+ SOLACE} (Ours)              & {0.71} & {0.67} & {0.288} & {5.39} & {22.41} & 0.278 & {0.87} & {3.11} \\

\midrule

SD3.5-L$\dagger$ (8.1B)  & 0.71 & 0.68 &   0.289 & 5.50 & 22.91 & 0.288 & 0.96 & 3.25 \\
(Reproduced)  & 0.51 & 0.68 &  0.284  & 5.28 & 21.86 & 0.264 & 0.70 & 2.98 \\
†\textbf{+ SOLACE} (Ours)              & 0.58 & 0.74 & 0.288 & 5.25 & 21.91 & 0.253 & 0.65 & 2.98 \\

\midrule

FLUX.1-Dev$\dagger$ (12B)               & 0.66 & 0.59 & 0.295 & 5.71 & 22.69 & 0.292  & 0.96  & 3.27 \\

(Reproduced) & 0.66 & 0.61 & 0.269 & 5.71 & 22.84 & 0.274  & 0.88  & 3.21 \\

\textbf{+ SOLACE} (Ours)              & 0.66 & 0.65 & 0.271 & 5.67 & 22.69 & 0.292 & 0.90 & 3.23 \\

\bottomrule
\end{tabular}
\caption{\textbf{Applying SOLACE to SD3.5-L~\cite{esser2024scaling} and FLUX.1-Dev~\cite{batifol2025flux}.} We apply SOLACE on additional models of SD3.5-L and FLUX.1-Dev, to verify the effect of SOLACE given (1) a larger base model, and (2) a different architecture from SD3.5-M.
$\dagger$ denotes results taken from DiffusionNFT~\cite{zheng2025diffusionnft}.
We base our experiments on our reproduced results based on the official weights of SD3.5-L~\cite{esser2024scaling} and FLUX.1-Dev~\cite{batifol2025flux}.
The results show that SOLACE consistently results in improved compositionality, text rendering and text-image alignment, while being competitive at human preference metrics.}
\label{tab:supp_other_models}
\end{table*}


\section{Applying SOLACE on SDXL}
\label{appendix:solace_sdxl}

To verify that SOLACE is not inherently tailored to DiT-based or flow-matching architectures, we apply SOLACE to \textbf{SDXL}~\cite{podell2023sdxl}, a UNet-based latent diffusion model.
Despite the architectural differences (SDXL uses a UNet backbone with DDPM-style noise scheduling rather than DiT-based flow matching), SOLACE produces consistent improvements in compositional generation (GenEval) and text rendering (OCR), as shown in \cref{tab:supp_sdxl}.
These results suggest that SOLACE's self-confidence reward is architecture-agnostic and can benefit UNet-based diffusion models as well.

\begin{table*}[h]
\centering
\small
\begin{tabular}{l c c c c c c c c}
\toprule
 & \multicolumn{2}{c}{Task-specific} & \multicolumn{2}{c}{Image Quality} & \multicolumn{4}{c}{Human Preference} \\
\cmidrule(lr){2-3}\cmidrule(lr){4-5} \cmidrule(lr){6-9}
  & GenEval & OCR  & ClipScore & Aesthetic & PickScore & HPSv2.1 & ImageReward & UnifiedReward\\
\midrule
SDXL~\cite{podell2023sdxl}   & 0.23  & 0.127 & 0.284 & 5.58 & 22.34 & 0.274 & 0.67 & 2.92 \\
\textbf{+ SOLACE} (Ours) & \textbf{0.25} & \textbf{0.144} & 0.284 & 5.57 & 22.33 & 0.270 & \textbf{0.70} & \textbf{2.94} \\
\bottomrule
\end{tabular}
\caption{\textbf{Applying SOLACE to SDXL~\cite{podell2023sdxl}.} SOLACE yields improvements in compositional generation and text rendering on a UNet-based diffusion model, demonstrating architecture-agnostic applicability.}
\label{tab:supp_sdxl}
\end{table*}



\section{SOLACE for Text-to-Video Generation}
\label{appendix:t2v}

\begin{figure*}[h]
\centering
\includegraphics[width=\linewidth]{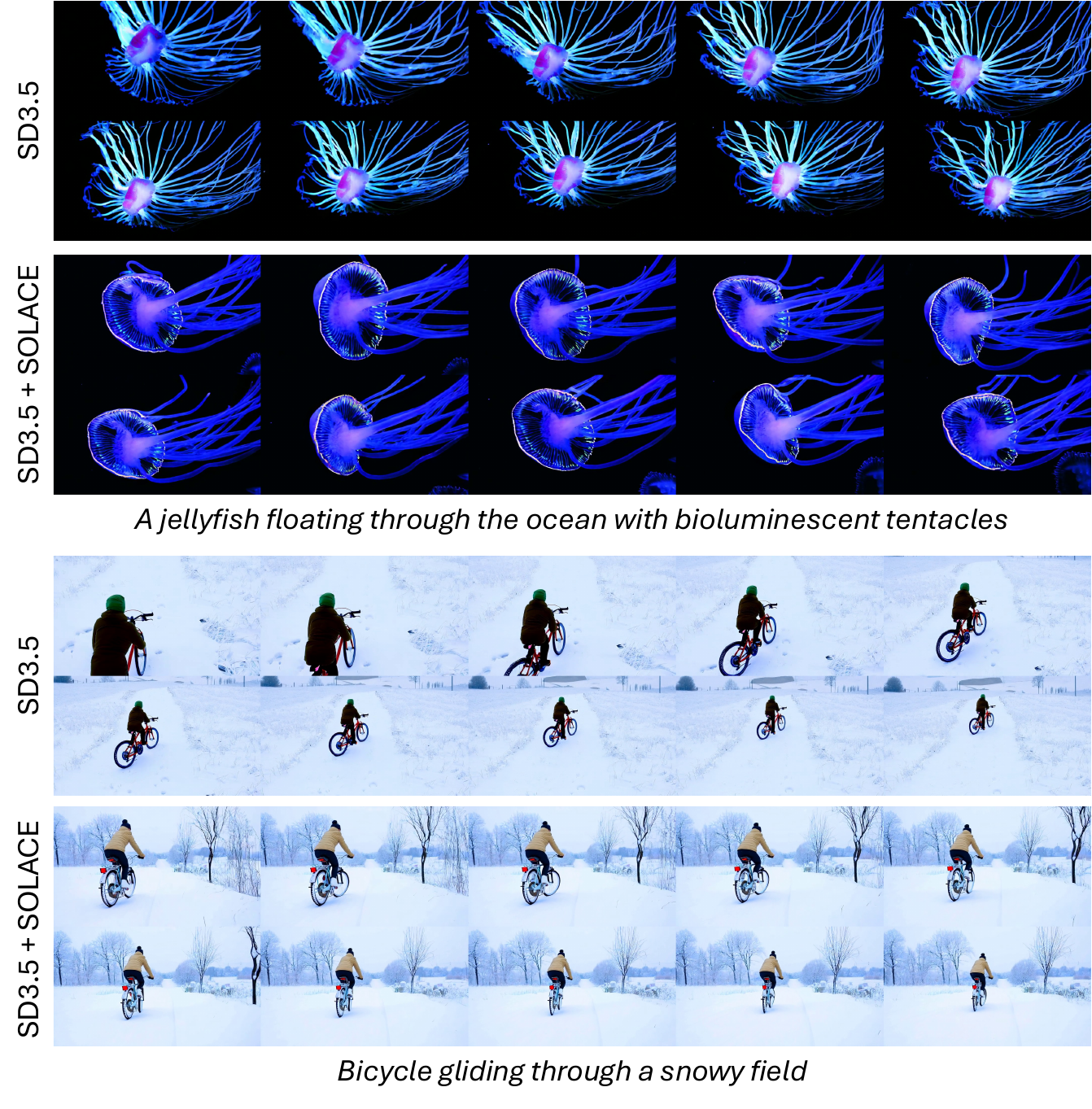}
\caption{\textbf{Qualitative results of SOLACE on Wan2.1-1.3B.} SOLACE produces videos with improved visual quality and prompt adherence compared to the base model.}
\label{fig:wan_solace_qual}
\end{figure*}

To test the applicability of SOLACE beyond text-to-image generation, we apply SOLACE to \textbf{Wan2.1-1.3B}~\cite{wan2025wan}, a text-to-video diffusion model.
We evaluate on the VBench-1.0~\cite{huang2024vbench} subset, and report the results in \cref{tab:t2v_vbench}.
As shown in the table, SOLACE yields improvements in subject consistency, background consistency, and dynamic degree, while maintaining competitive motion smoothness, demonstrating that SOLACE generalizes effectively to the text-to-video generation setting.
Qualitative results are provided in \cref{fig:wan_solace_qual}.
For instance, in the jellyfish example (top), SOLACE produces noticeably more stable jellyfish movements compared to the baseline.
In the ``bicycle gliding through a snowy field'' example (bottom), the baseline generates an unnatural gliding motion where the gliding direction does not match the bicycle's orientation, whereas SOLACE produces a much more natural and coherent gliding motion.

\begin{table}[h]
\centering
\footnotesize
\setlength{\tabcolsep}{3pt}
\begin{tabular}{l c c c c c}
\toprule
 & Subj. & BG & Aesth. & Motion & Dyn. \\
 & Consist. & Consist. & Qual. & Smooth. & Deg. \\
\midrule
Wan2.1-1.3B & 0.94 & 0.96 & \textbf{0.59} & \textbf{0.97} & 0.47 \\
\textbf{+ SOLACE} & \textbf{0.95} & \textbf{0.97} & 0.58 & \textbf{0.97} & \textbf{0.51} \\
\bottomrule
\end{tabular}
\caption{\textbf{Applying SOLACE to Wan2.1-1.3B for text-to-video generation.} Evaluation on VBench-1.0 subset. SOLACE improves subject consistency, background consistency, and dynamic degree while maintaining competitive motion smoothness.}
\label{tab:t2v_vbench}
\end{table}

\section{Resolution Analysis}
\label{appendix:resolution}

Our main experiments use $512\times512$ resolution for both training and evaluation, following the configuration of Flow-GRPO~\cite{liu2025flowgrpo}.
To verify that the improvements transfer across resolutions, we additionally train SOLACE at $1024\times1024$ resolution and evaluate both models at both scales.

As shown in \cref{tab:resolution}, SOLACE trained at $512\times512$ (SOLACE$_{512}$) transfers well to $1024\times1024$ inference, yielding consistent improvements in GenEval and OCR at the higher resolution.
SOLACE trained directly at $1024\times1024$ (SOLACE$_{1024}$) also shows gains, though with a slightly different trade-off profile across metrics.
These results confirm that SOLACE's benefits are not resolution-specific.

\begin{table*}[h]
\centering
\small
\begin{tabular}{l c c c c c c c c}
\toprule
 & \multicolumn{2}{c}{Task-specific} & \multicolumn{2}{c}{Image Quality} & \multicolumn{4}{c}{Human Preference} \\
\cmidrule(lr){2-3}\cmidrule(lr){4-5} \cmidrule(lr){6-9}
  & GenEval & OCR  & ClipScore & Aesthetic & PickScore & HPSv2.1 & ImageReward & UnifiedReward\\
\midrule
\multicolumn{9}{l}{\textit{Inference at $512\times512$}} \\
\midrule
SD3.5-M  & 0.65  & 0.61 & 0.282 & 5.36 & 22.34 & 0.279 & 0.84 & 3.08 \\
+ SOLACE$_{512}$  & \textbf{0.71}  & \textbf{0.67} & \textbf{0.288} & \textbf{5.39} & \textbf{22.41} & \textbf{0.284} & \textbf{0.87} & \textbf{3.10} \\
+ SOLACE$_{1024}$  & 0.68  & 0.63 & 0.284 & \textbf{5.39} & 22.39 & \textbf{0.284} & \textbf{0.87} & \textbf{3.10} \\
\midrule
\multicolumn{9}{l}{\textit{Inference at $1024\times1024$}} \\
\midrule
SD3.5-M  & 0.65  & 0.57 & \textbf{0.293} & \textbf{5.98} & 21.91 & \textbf{0.305} & \textbf{1.15} & \textbf{3.48} \\
+ SOLACE$_{512}$  & \textbf{0.71}  & \textbf{0.64} & 0.292 & 5.95 & 21.68 & 0.283 & 1.00 & 3.41 \\
+ SOLACE$_{1024}$  & 0.68  & 0.63 & 0.289 & 5.38 & \textbf{22.48} & 0.283 & 0.93 & 3.19 \\
\bottomrule
\end{tabular}
\caption{\textbf{Resolution analysis.} SOLACE trained at $512\times512$ transfers effectively to $1024\times1024$ inference, with consistent gains in compositional generation and text rendering across resolutions.}
\label{tab:resolution}
\end{table*}



\section{Comparison with Closed-Source Models}
\label{appendix:closed_source}

To contextualize SOLACE's improvements, we evaluate two closed-source models (Gemini 2.5-Flash and GPT-image-1.5) on our benchmark suite.
As shown in \cref{tab:closed_source}, closed-source models achieve higher absolute scores due to larger model capacities and proprietary training data. Nevertheless, SOLACE narrows the gap from the SD3.5-M baseline, particularly in compositional generation and text rendering.

\begin{table*}[h]
\centering
\small
\begin{tabular}{l c c c c c c c c}
\toprule
 & \multicolumn{2}{c}{Task-specific} & \multicolumn{2}{c}{Image Quality} & \multicolumn{4}{c}{Human Preference} \\
\cmidrule(lr){2-3}\cmidrule(lr){4-5} \cmidrule(lr){6-9}
  & GenEval & OCR  & ClipScore & Aesthetic & PickScore & HPSv2.1 & ImageReward & UnifiedReward\\
\midrule
SD3.5-M  & 0.65  & 0.61 & 0.282 & 5.36 & 22.34 & 0.279 & 0.84 & 3.08 \\
\textbf{+ SOLACE} (Ours) & 0.71  & 0.67 & 0.288 & 5.39 & 22.41 & 0.278 & 0.87 & 3.11 \\
\midrule
Gemini 2.5-Flash  & 0.75 & 0.72 & 0.270 & 5.70 & 23.02 & 0.287 & 0.79 & 3.45 \\
GPT-image-1.5 & 0.84  & 0.81 & 0.286 & 5.54 & 23.24 & 0.301 & 1.11 & 3.57 \\
\bottomrule
\end{tabular}
\caption{\textbf{Comparison with closed-source models.} While closed-source models achieve higher absolute scores due to larger capacities and proprietary training, SOLACE narrows the gap from the SD3.5-M baseline, particularly in compositional generation and text rendering.}
\label{tab:closed_source}
\end{table*}



\section{Training Collapse Analysis}
\label{appendix:training_collapse_analysis}

\noindent
\textbf{When and why collapse occurs.}
We monitor the batch-mean self-confidence (negative log error, averaged over probes and probed timesteps) across training iterations.
Collapse is characterized by a rapid, sustained surge in this score (an overconfidence spike), followed by degenerate, low-texture generations (reward hacking).
Empirically, two settings precipitate this behavior: (i) training on too many timesteps (\(\rho>0.6\) in \(|\mathcal{T}_{\mathrm{train}}|=\lceil \rho\,|\mathcal{T}|\rceil\)), which exposes early, easily exploitable steps; and (ii) sampling the \(G\) rollout candidates \emph{without} CFG, which reduces exploration and inflates apparent self-confidence.
A KL anchor alone is insufficient to prevent these modes.

\noindent
\textbf{Mitigations used in SOLACE.}
We restrict training to the latter \(60\%\) of steps (\(\rho=0.6\)), keep CFG \emph{on} during rollouts (but \emph{off} when scoring self-confidence), and retain clipping, per-timestep weighting, and antithetic probes.
These choices suppress overconfidence spikes and stabilize learning.

\noindent
\textbf{Why SOLACE's reward is amenable to targeted stabilization.}
Since the reward is a monotonic transform of denoising error (\ie $r=-\log(\mathrm{MSE}+\delta)$), the degenerate solution is concrete and diagnosable: maximizing $\mathbb{E}_{z_0\sim\pi_\theta(\cdot|c)}[r(z_0)]$ can steer samples toward latent regimes where injected noise becomes trivially predictable (\eg low-variance, textureless outputs).
Because self-confidence is not a fixed black-box oracle, we can directly modify the \emph{reward computation itself} (solver-aligned timestep probing, suffix-window training, and no-CFG scoring) to suppress these shortcut solutions, rather than relying solely on generic stabilizers (\eg KL weights or reward scaling) that do not change what the reward measures.

\input{assets/figures/model_collapse}



\section{Diversity and Semantic Correctness Analysis}
\label{appendix:diversity_semantic}

A natural concern with self-confidence as a reward is whether it biases the model toward high-density but semantically incorrect modes, or reduces sample diversity.
We address both concerns empirically.

\smallbreak
\noindent
\textbf{Semantic correctness on rare compositions.}
Self-confidence is computed \emph{under the same text conditioning} $c$ in $r(x,c)$, which reduces pressure toward prompt-agnostic high-density modes.
To test whether SOLACE degrades on less common compositional prompts, we evaluate on RareBench~\cite{park2025rarebench}, a benchmark consisting of diverse and complex rare concept compositions.
As shown in \cref{tab:diversity_semantic}, CLIPScore on RareBench is largely preserved after SOLACE post-training, suggesting no measurable degradation on rare or out-of-distribution compositions.

\smallbreak
\noindent
\textbf{Diversity preservation.}
We measure diversity using the mean pairwise CLIP embedding distance across 64 samples per prompt on 50 DrawBench~\cite{saharia2022photorealistic} prompts.
As reported in \cref{tab:diversity_semantic}, the diversity score is maintained (and even slightly improved) after SOLACE post-training.
This is consistent with the fact that SOLACE's reward measures conditional denoising self-consistency rather than explicitly minimizing conditional entropy $H(x|c)$, and the GRPO objective with KL regularization provides sufficient diversity preservation.

\begin{table}[h]
\centering
\small
\begin{tabular}{l c c}
\toprule
 & CLIPScore$\uparrow$ & Diversity Score$\uparrow$ \\
 & (RareBench) & (DrawBench) \\
\midrule
SD3.5-M & 0.2752 & 0.9519 \\
SD3.5-M + SOLACE & 0.2746 & 0.9545 \\
\bottomrule
\end{tabular}
\caption{\textbf{Semantic correctness and diversity analysis.} CLIPScore on RareBench~\cite{park2025rarebench} (rare compositions) and diversity score on DrawBench~\cite{saharia2022photorealistic} (64 samples per prompt, 50 prompts) show that SOLACE preserves both semantic accuracy on uncommon concepts and sample diversity.}
\label{tab:diversity_semantic}
\end{table}



\section{Effect of Negative Advantages}
\label{appendix:positive_only}

SOLACE uses GRPO, where updates are weighted by a signed, within-group advantage: samples with below-average self-confidence receive negative advantages and are explicitly downweighted.
To verify the importance of this negative signal, we compare against a \emph{positive-only} variant that clips advantages to be non-negative (\ie $\max(\widehat{A}_t^{\,i}, 0)$), effectively removing the penalty for low-confidence samples.

As shown in \cref{tab:positive_only}, the positive-only variant underperforms the full SOLACE objective on GenEval, OCR, and CLIPScore, confirming that negative advantages provide important learning signal.
While the positive-only variant achieves higher aesthetic and some human preference scores, it sacrifices the core compositional and text-rendering gains that SOLACE targets.

\begin{table*}[h]
\centering
\small
\begin{tabular}{l c c c c c c c c}
\toprule
 & \multicolumn{2}{c}{Task-specific} & \multicolumn{2}{c}{Image Quality} & \multicolumn{4}{c}{Human Preference} \\
\cmidrule(lr){2-3}\cmidrule(lr){4-5} \cmidrule(lr){6-9}
  & GenEval & OCR  & ClipScore & Aesthetic & PickScore & HPSv2.1 & ImageReward & UnifiedReward\\
\midrule
SD3.5-M  & 0.65  & 0.61 & 0.282 & 5.36 & 22.34 & 0.279 & 0.84 & 3.08 \\
\textbf{+ SOLACE}  & \textbf{0.71}  & \textbf{0.67} & \textbf{0.288} & 5.39 & \textbf{22.41} & 0.278 & 0.87 & 3.11 \\
+ SOLACE (positive-only) & 0.69  & 0.62 & 0.285 & \textbf{5.80} & 21.57 & \textbf{0.281} & \textbf{0.91} & \textbf{3.20} \\
\bottomrule
\end{tabular}
\caption{\textbf{Effect of negative advantages.} Removing negative advantages (positive-only variant) degrades compositional generation, text rendering, and text-image alignment, demonstrating that the full signed advantage is important for SOLACE's effectiveness.}
\label{tab:positive_only}
\end{table*}



\section{Additional Ablation Studies}
\label{appendix:ablation_study}

We conduct additional ablation studies and comparative experiments to validate the design choices of SOLACE. The results are summarized in \cref{tab:supp_ablation}.

\subsection{Caption datasets for SOLACE}
\label{app:ablation_captions}
SOLACE relies on intrinsic self-confidence and thus requires only prompts (not external reward models). We compare three prompt sources: (i) \emph{text-rendering (OCR)} prompts from Flow-GRPO~\cite{liu2025flowgrpo} (our default), (ii) \emph{PickScore}~\cite{kirstain2023pick} prompts, and (iii) \emph{GenEval}~\cite{ghosh2023geneval} prompts. As shown in~\cref{tab:supp_other_models}, denser, more prescriptive prompts (OCR) yield the strongest gains; empirically, self-confidence is most reliable when the text condition is explicit and descriptive.
We provide the descriptions and examples for each prompt dataset in~\cref{tab:solace_captions}.

\begin{table*}[h]
\centering
\small
\begin{tabularx}{\linewidth}{lX}
\toprule
\multicolumn{2}{l}{\textbf{(i) Text-rendering (OCR) prompts — default}}\\
\cmidrule(lr){1-2}
\textit{Characteristics} & Dense, explicit textual content (exact strings, font/placement hints), strong conditioning for legibility and alignment.\\
\textit{Examples} & ``A postage stamp design featuring the motto "Unity in Diversity", showcasing a vibrant collage of people from various ethnic backgrounds, each holding hands in a circle, set against a backdrop of colorful, interwoven patterns symbolizing unity and cultural richness.''\\
 & ``In a luxurious hotel lobby, an elegant digital display above the elevator reads "Now Playing". Soft, ambient elevator music fills the space, enhancing the serene and welcoming atmosphere. A plush, modern sofa and a glass coffee table are seen in the foreground, with polished marble floors reflecting the ambient light.''\\
  & ``A sleek, modern corporate lobby featuring a large, minimalist sculpture prominently inscribed with "Innovate or Perish", reflecting the company's commitment to forward-thinking. The sculpture stands against a backdrop of glass and steel, with subtle lighting enhancing its form and the powerful message it conveys.''\\
\addlinespace[4pt]
\midrule
\multicolumn{2}{l}{\textbf{(ii) PickScore prompts}}\\
\cmidrule(lr){1-2}
\textit{Characteristics} & Open-ended statements; often adds context with simple concatentation of adjectives; weaker constraints on text content/layout.\\
\textit{Examples} & ``An attractive young woman petting a cat''\\
 & ``(a girl in steampunk fantasy world), (ultra detailed prosthetic arm and leg), (beautifully drawn face:1.2), blueprints, (magic potions:1.4), mechanical tools,  plants, (a small cat:1.1), silver hair, (full body:1.2), magic dust, books BREAK (complex ultra detailed of medieval fantasy city), (steampunk fantasy:1.2), indoors, workshop, (Steam-powered machines:1.2), (clockwork automatons:1.2), (a small wooden toy), (intricate details:1.6),  lamps, colorful details, iridescent colors, BREAK illustration, ((masterpiece:1.2, best quality)), 4k, ultra detailed, solo, (photorealistic:1.2), asymmetry, looking at viewer, smile''\\
 & ``Cyborg cow, cyberpunk alien india, body painting, bull, star wars design, third eye, mehendi body art, yantra, cyberpunk mask, baroque style, dark fantasy, kathakali characters, high tech, detailed, spotlight, shadow color, high contrast, cyberpunk city, neon light, colorful, bright, high tech, high contrast, synthesized body, hyper realistic, 8k, epic ambient light, octane rendering, kathakali, soft ambient light, HD,''\\
\addlinespace[4pt]
\multicolumn{2}{l}{\textbf{(iii) GenEval prompts}}\\
\cmidrule(lr){1-2}
\textit{Characteristics} & Compositional verification (objects, counts, relations), moderate specificity, minimal typography.\\
\textit{Examples} & ``a photo of a yellow bus and an orange handbag''\\
 & ``a photo of four surfboards''\\
  & ``a photo of a book left of a cat''\\
\bottomrule
\end{tabularx}
\caption{\textbf{Prompt sources compared for SOLACE.} Denser, text-focused prompts (OCR) provide stronger supervision signals for intrinsic self-confidence, leading to larger gains than more open-ended (PickScore) or simple compositional (GenEval) prompts.}
\label{tab:solace_captions}
\end{table*}

\subsection{Effect of group size}
\label{app:ablation_group}
We clarify a typographical error in the main paper: although we stated $G{=}24$, all experiments used $G{=}16$. Varying $G$ shows that $G{=}16$ outperforms $G{=}8$ (more within-prompt exploration improves group-relative normalization) while $G{=}32$ destabilizes training: larger groups reduce the number of distinct prompts per batch, lowering inter-prompt diversity and increasing the risk of over-optimization under relative advantages. In practice, $G{=}16$ strikes a robust compute–stability trade-off.

\subsection{Stepwise vs.\ aggregated reward}
\label{app:ablation_stepwise}
Although SOLACE's self-confidence can be computed per step, we find that using the \emph{aggregated} reward, \ie, averaging weighted per-step scores over the probed timesteps, consistently performs better than optimizing stepwise advantages.
Stepwise improvements at individual timesteps need not translate to a better final sample and tend to increase variance and solver sensitivity; aggregation provides a more stable, outcome-aligned signal for post-training.

\begin{table*}[!ht]

\centering
\small
\begin{tabular}{l c c c c c c c c}
\toprule
 & \multicolumn{2}{c}{Task-specific} & \multicolumn{2}{c}{Image Quality} & \multicolumn{4}{c}{Human Preference} \\
\cmidrule(lr){2-3}\cmidrule(lr){4-5} \cmidrule(lr){6-9}

  & GenEval & OCR  & ClipScore & Aesthetic & PickScore & HPSv2.1 & ImageReward & UnifiedReward\\
\midrule
\multicolumn{9}{l}{\textit{Caption dataset used for SOLACE}} \\
\midrule
-  & 0.65 & 0.61 & 0.282 & 5.36 & 22.34 & 0.279  & 0.84  & 3.08 \\

PickScore prompts             & 0.70 & 0.62 & 0.285 & 5.26 & 22.13 & 0.278 & 0.65 & 2.96 \\

GenEval prompts              & 0.71 & 0.62 & 0.286 & 5.32 & 22.35 & 0.275 & 0.80 & 3.05 \\

OCR prompts (Ours)              & \textbf{0.71} & \textbf{0.67} & \textbf{0.288} & \textbf{5.39} & \textbf{22.41} & 0.278 & \textbf{0.87} & \textbf{3.11} \\

\midrule
\multicolumn{9}{l}{\textit{Group size $G$}} \\
\midrule

8              & 0.70 & 0.64 & 0.285 & 5.29 & 22.28 & 0.267 & 0.75 & 3.00 \\
16 (Ours)              & \textbf{0.71} & \textbf{0.67} & \textbf{0.288} & \textbf{5.39} & \textbf{22.41} & 0.278 & \textbf{0.87} & \textbf{3.11} \\
32              & 0.61 & 0.51 & 0.274 & 5.18 & 21.73 & 0.226 & 0.16 & 2.73 \\

\midrule
\multicolumn{9}{l}{\textit{Step-wise reward vs. Aggregated reward}} \\
\midrule

Stepwise              & 0.67 & 0.60 & 0.285 & \textbf{5.39} & 22.36 & 0.277 & 0.83 & 3.07 \\

Aggregated (Ours)              & \textbf{0.71} & \textbf{0.67} & \textbf{0.288} & \textbf{5.39} & \textbf{22.41} & 0.278 & \textbf{0.87} & \textbf{3.11} \\

\bottomrule
\end{tabular}
\caption{\textbf{Additional ablation/comparative results.} The results show that our current design choices for the (1) Caption dataset used, (2) Group size $G$, and (3) Aggregated self-confidence rewrads yield the best performances.}
\label{tab:supp_ablation}
\end{table*}



\section{Additional Implementation Details}
\label{appendix:implementation_details}

In this section we summarize the main implementation choices used in our SOLACE training pipeline.
\textbf{We acknowledge and correct a typographical error in the main paper}: although we stated that the group size was $G = 24$, all experiments were in fact conducted with $G = 16$.
The summary of hyperparameters and configurations is illustrated in \cref{tab:solace_hparams_single}.
\begin{table*}[t]
\centering
\small
\renewcommand{\arraystretch}{1.15}
\begin{tabular}{lll}
\toprule
\textbf{Category} & \textbf{Hyperparameter} & \textbf{Value (SOLACE, SD3.5-M)} \\
\midrule
Model & Base model & \texttt{stabilityai/stable-diffusion-3.5-medium} (SD3.5-M) \\
      & Components trained & Transformer (denoiser) only; VAE and all text encoders frozen \\
\midrule
LoRA  & LoRA usage & \texttt{use\_lora = True} \\
      & Rank $r$ & $32$ \\
      & Scaling factor $\alpha$ & $64$ \\
      & Init of LoRA weights & Gaussian \\
      & Target modules & \texttt{attn.add\_k\_proj}, \texttt{attn.add\_q\_proj}, \\ & & \texttt{attn.add\_v\_proj}, \texttt{attn.to\_add\_out}, \\
      &                  & \texttt{attn.to\_k}, \texttt{attn.to\_q}, \texttt{attn.to\_v}, \texttt{attn.to\_out.0} \\
\midrule
Data / prompts & Train / test files & \texttt{train.txt}, \texttt{test.txt} (one prompt per line) \\
               & Tokenization & SD3.5 tokenizers; max length 128 (embeddings), 256 (logging) \\
\midrule
Sampling & Image resolution & $512 \times 512$ \\
         & Sampler steps (train / eval) & train:10, eval:40 \\
         & Train timestep fraction & \texttt{train.timestep\_fraction = 0.99} $\Rightarrow T_{\mathrm{train}} = 9$ \\
         & Suffix proportion $\rho$ in GRPO & 0.6 \\
         & Guidance scale (train/eval) & \texttt{sample.guidance\_scale = 4.5} \\
         & Noise level (SDE step) & \texttt{sample.noise\_level = 0.7} \\
         & Train batch size / GPU (sampling) & \texttt{sample.train\_batch\_size = 8} images \\
         & Test batch size / GPU & \texttt{sample.test\_batch\_size = 16} images \\
         & Images per prompt (group size $G$) & \texttt{sample.num\_image\_per\_prompt = 16} \\
         & Number of GPUs & $8$ \\
         & Batches per epoch (sampling) & \texttt{sample.num\_batches\_per\_epoch = 4} \\
         & Global samples / batch & $8$ (bs) $\times\, 8$ (GPUs) $= 64$ images \\
         & Prompts / batch & $64 / 16 = 4$ prompts per sampling batch \\
         & Same latent per prompt & \texttt{sample.same\_latent = False} \\
\midrule
Self-confidence (SOLACE) & Probes per step $K$ & $8$ (antithetic pairing: $K/2$ noise, $K/2$ negated) \\
                      & Probe timesteps & Last half of used timesteps: $j = 4,\dots,8$ (for $T_{\mathrm{train}}=9$) \\
                      & Noise schedule for probe & $\lambda_t = \tau_t / 1000$; $x_t = (1-\lambda_t)x_0 + \lambda_t \epsilon$ \\
                      & Per-step score & $s_t = -\log(\mathrm{MSE}_t + 10^{-6})$, MSE between injected and predicted noise \\
                      & Normalization & Per-timestep batch-wise z-score, then mean over timesteps \\
                      & CFG inside probe & Disabled (conditional branch only) \\
\midrule
Training (GRPO)
                & PPO / GRPO clip range & $\rho_{i,t}$ clipped to $[1-\texttt{clip\_range},\,1+\texttt{clip\_range}]$ (PPO style) \\
                & KL regularizer weight & \texttt{train.beta = 0.04} \\
                & KL form & $D_{\mathrm{KL}} = \|\mu_\theta - \mu_{\mathrm{ref}}\|_2^2 \big/ (2 \sigma_t^2)$ (mean-only Gaussian) \\

\midrule
Optimization / EMA & Optimizer & AdamW on LoRA parameters (no base-parameter updates) \\
                   & Learning rate & $3\times 10^{-4}$ (constant) \\
                   & Gradient clipping & Global norm clipping at \texttt{train.max\_grad\_norm} \\
                   & EMA usage & \texttt{train.ema = True} \\
                   & EMA decay & $0.9$ \\
                   & EMA update interval & Every $8$ optimizer steps (\texttt{update\_step\_interval = 8}) \\
                   & EMA usage in eval & EMA weights used for evaluation; online weights restored afterwards \\
\midrule
External rewards / eval & Training reward & Internal self-confidence only (no external reward in training) \\
                        & SDS-only eval & Optional SDS self-confidence evaluation on EMA model for monitoring \\
\bottomrule
\end{tabular}
\caption{Hyperparameters and key implementation details for SOLACE training on SD3.5-M.}
\label{tab:solace_hparams_single}
\end{table*}

\subsection{Base models and LoRA configuration}

We build on the \texttt{StableDiffusion3Pipeline} from \texttt{diffusers} with the pretrained model SD3.5-M: \texttt{stabilityai/stable-diffusion-3.5-medium}.
We freeze all components except the denoiser: the VAE and all text encoders are kept fixed and used only for inference. Only the main transformer (UNet-like denoiser) is updated during training, based on LoRA.
We run the text encoders in mixed precision (\texttt{fp16} in our main SOLACE runs) and keep the VAE in \texttt{fp32} for stability.

For parameter-efficient fine-tuning we apply LoRA to the transformer with
\begin{itemize}
    \item LoRA rank $r = 32$ and scaling factor $\alpha = 64$,
    \item Gaussian initialization of LoRA weights,
    \item Target modules inside each attention block:
    \begin{center}
        \texttt{attn.add\_k\_proj}, \texttt{attn.add\_q\_proj}, \texttt{attn.add\_v\_proj}, \texttt{attn.to\_add\_out},\\
        \texttt{attn.to\_k}, \texttt{attn.to\_q}, \texttt{attn.to\_v}, \texttt{attn.to\_out.0}.
    \end{center}
\end{itemize}
All non-LoRA base weights remain frozen.

\subsection{Datasets and prompt processing}

We consider two kinds of prompt datasets:
\begin{itemize}
    \item \textbf{Plain text prompt datasets.}
    We store the prompts in plain text files \texttt{train.txt} and \texttt{test.txt}. Each line contains a single prompt string. (\eg PickScore, Text Rendering dataset)
    \item \textbf{GenEval-style metadata.}
    For experiments on GenEval-style prompts we use JSONL files \texttt{\{train,test\}\_metadata.jsonl}, where each line is a JSON object that contains at least a \texttt{"prompt"} field and additional metadata.
\end{itemize}

For each batch of prompts we compute text embeddings using the three SD3.5 text encoders. We also precompute embeddings for the empty prompt \texttt{""} and use them as unconditional embeddings for classifier-free guidance (CFG) during sampling and log-probability computation.

\subsection{Distributed sampling and grouping}

We use HuggingFace Accelerate~\cite{hfaccelerate} for distributed training. Let $N$ be the number of GPUs (processes), and let $B_{\text{sample}}$ denote the per-device {sample batch size}.
In our main SOLACE setting we use $N = 8, B_{\text{sample}} = 8, G = 16$.
Thus a single sampling batch contains $N B_{\text{sample}} = 64$ images, corresponding to $64 / 16 = 4$ distinct prompts, each with $G=16$ candidate images.
We train for 2,000 iterations, which takes around 30 hours on 8$\times$NVIDIA 332
RTX PRO 6000 Blackwell GPUs.

\subsection{KL regularization}
Following Flow-GRPO, regularize the policy via a KL term that constrains the transition mean to stay close to a reference (the base model without LoRA):
\begin{itemize}
    \item The SDE step module returns the current mean \(\mu_\theta\) and a reference variance \(\sigma_t^2\).
    \item We compute a reference mean \(\mu_{\text{ref}}\) by temporarily disabling LoRA adapters and re-evaluating the same step.
    \item Assuming Gaussian transitions with equal variance, the per-step KL divergence simplifies to
    \[
        D_{\mathrm{KL}} = \frac{1}{2 \sigma_t^2} \left\| \mu_\theta - \mu_{\text{ref}} \right\|_2^2.
    \]
\end{itemize}
We average this KL over spatial dimensions and the batch and add it to the policy loss with weight \(\beta = 0.04\).


\section{User Study Instructions and Interface}
\label{appendix:user_study_interface}

We provide the details of the instructions and interface used for the user study.

\paragraph{Instructions.}
For each text prompt, you will be shown a pair of \emph{AI-generated} images (left and right).
For every image pair, you are asked to answer the following two questions \emph{independently}:

\begin{enumerate}
    \item \textbf{Visual realism and appeal:}
    Which image do you find to be more visually realistic and appealing?

    \item \textbf{Text–image alignment:}
    Which image better aligns with the given text description?
\end{enumerate}

For each question, please select your preferred image (left or right) based solely on the specified criterion.

\begin{figure}[t]
    \centering
    \includegraphics[width=\linewidth]{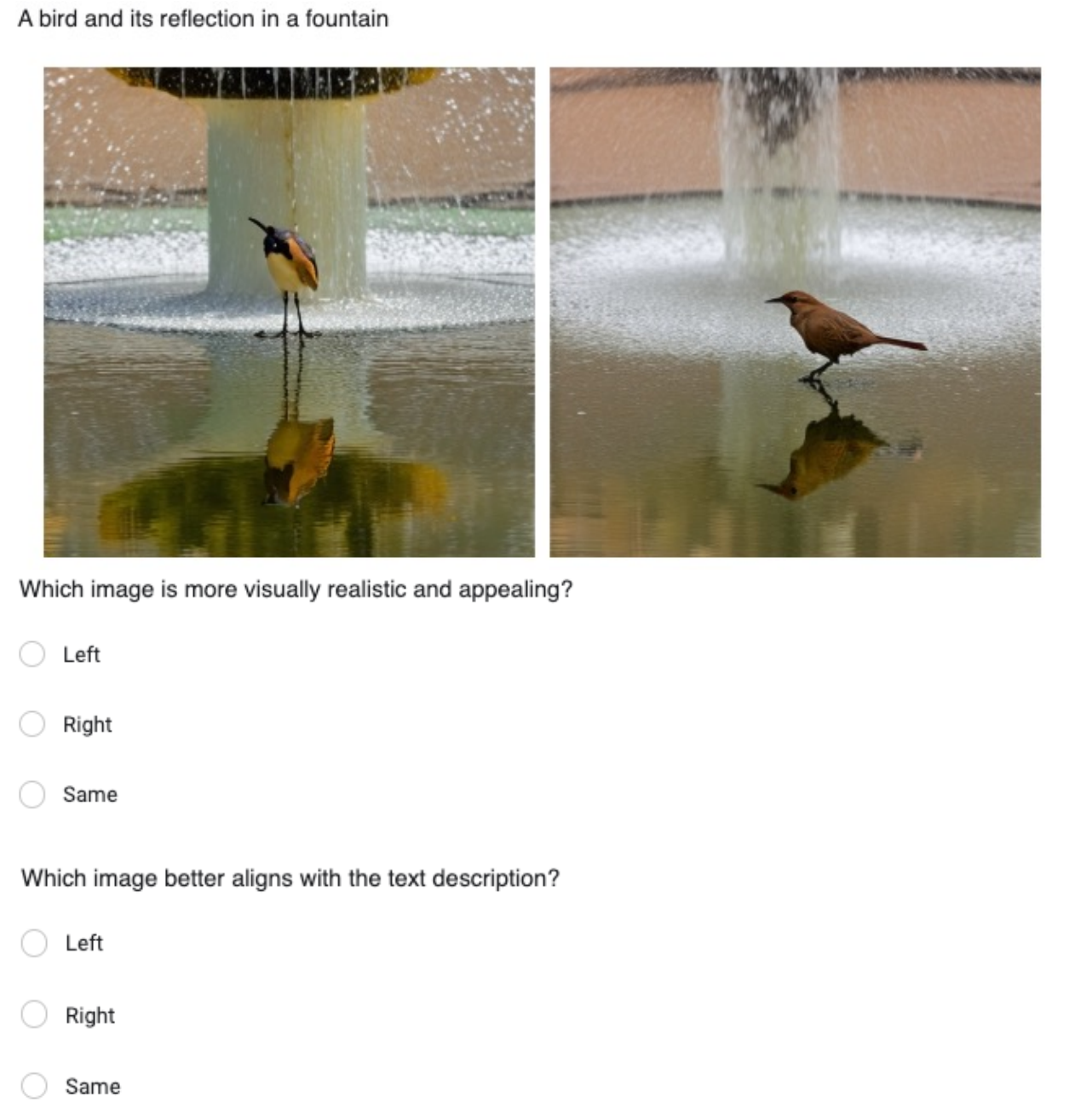}%
    \caption{User study interface used to collect human preferences between pairs of AI-generated images.}
    \label{fig:user_study_interface}
\end{figure}

\paragraph{Interface.}
The user interface used in the study is illustrated in \cref{fig:user_study_interface}.

\section{Additional Qualitative Results}
\label{appendix:additional_qual}

We provide side-by-side samples for (i) PickScore–post-trained (Flow-GRPO) SD3.5–M, (ii) FLUX.1–Dev, and (iii) SD3.5–L in \cref{fig:supp_main_qual}.
Across diverse prompts, SOLACE yields visibly sharper text rendering, more faithful object counts and relations, and fewer artifacts, echoing the quantitative gains in compositionality, text rendering, and text–image alignment, with no obvious regressions on non-target aspects.

\input{assets/figures/supp_main_qual}

%% file: assets/figures/model_collapse.tex
\begin{figure}[h]
\begin{center}
\includegraphics[width=1.0\linewidth]{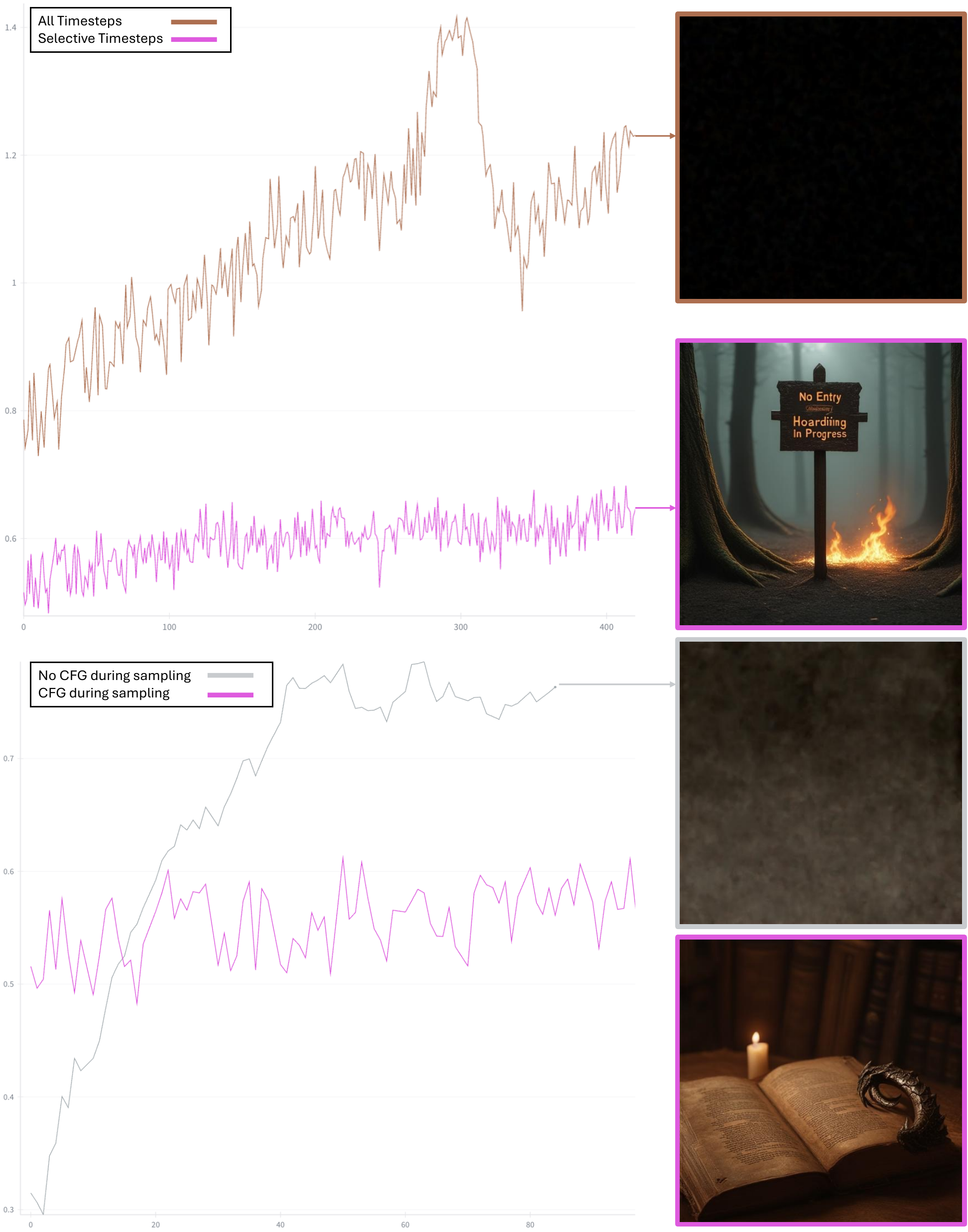}
\end{center}
\caption{\textbf{Visualization of training collapse in SOLACE.} Self-confidence (y-axis) versus training iteration under different settings. 
Using \(\rho>0.6\) or sampling rollouts without CFG drives a steep, short-horizon increase in self-confidence, followed by degenerate outputs—evidence of reward hacking. 
SOLACE’s default settings (\(\rho{=}0.6\) and CFG for rollouts) avoid this behavior while preserving steady improvements.}
\label{fig:model_collapse}
\end{figure}

%% file: assets/figures/supp_main_qual.tex
\begin{figure*}[h]
\begin{center}
\includegraphics[width=1.0\linewidth]{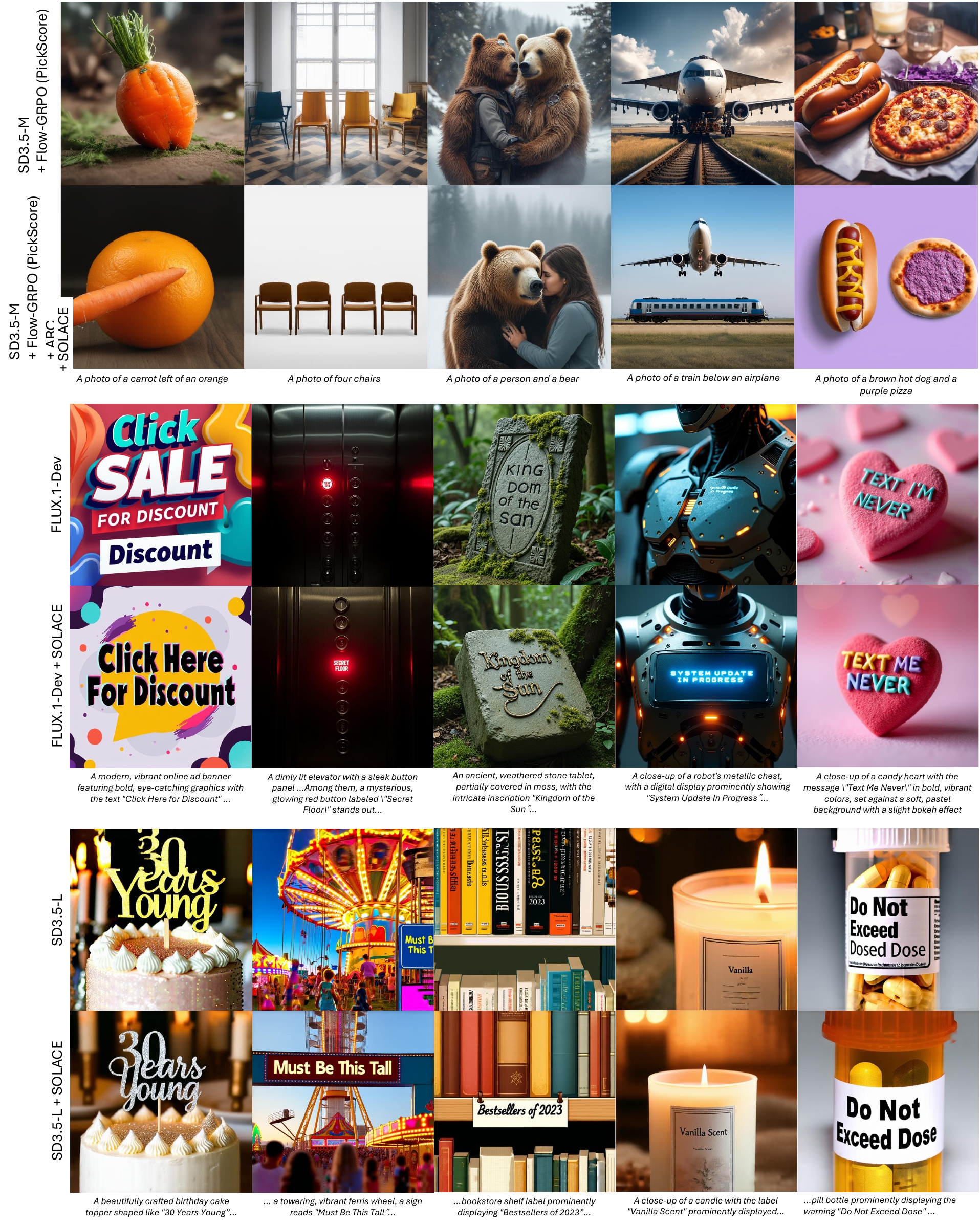}
\end{center}
\vspace{-5mm}
\caption{\textbf{Additional qualitative results of SOLACE.} We present additional qualitative results of SOLACE when applied to (1) Flow-GRPO~\cite{liu2025flowgrpo} post-trained SD3.5-M~\cite{esser2024scaling}, (2) FLUX.1-Dev~\cite{batifol2025flux}, and (3) SD3.5-L~\cite{esser2024scaling}. Best viewed on electronics.}
\label{fig:supp_main_qual}
\end{figure*}